\documentclass[11pt]{article}

\usepackage[margin=1in]{geometry}
\usepackage{amsmath,amssymb,amsthm}
\usepackage{booktabs}
\usepackage{tabularx}
\usepackage{hyperref}
\usepackage[table]{xcolor}
\hypersetup{colorlinks=true, linkcolor=blue!50!black, citecolor=blue!50!black, urlcolor=blue!50!black}
\usepackage{tikz}
\usetikzlibrary{positioning,arrows.meta,shapes.geometric,fit,calc,backgrounds,matrix,decorations.pathreplacing}
\usepackage{enumitem}
\setlist[itemize]{leftmargin=*,labelindent=0pt,itemsep=1pt,topsep=3pt,parsep=0pt,partopsep=0pt}
\usepackage{microtype}
\usepackage{graphicx}
\usepackage{subcaption}
\usepackage{pifont}
\usepackage{fontawesome5}
\usepackage[section]{placeins}
\usepackage{float}

\usepackage{fontspec}
\newcommand{\leanPathwiseDifferentiableAt}{\texttt{Pathwise\allowbreak Differentiable\allowbreak At}}
\newcommand{\leanPathwiseDifferentiableAtVec}{\texttt{Pathwise\allowbreak Differentiable\allowbreak At\_vec}}

\theoremstyle{definition}
\newtheorem{theorem}{Theorem}[section]
\newtheoremstyle{noparens}{}{}{}{}{\bfseries}{.}{ }{\thmname{#1}\thmnote{ #3}}
\theoremstyle{noparens}

\theoremstyle{definition}

\newtheorem{definition}[theorem]{Definition}

\usepackage[most]{tcolorbox}
\tcolorboxenvironment{theorem}{
  enhanced jigsaw,
  boxrule=0pt, frame hidden,
  boxsep=0pt, top=4pt, bottom=4pt, left=8pt, right=8pt,
  arc=0pt, outer arc=0pt,
  colback=blue!4!white,
  borderline west={1.2pt}{0pt}{blue!50!cyan!60},
  before skip=8pt, after skip=8pt, breakable,
}
\tcolorboxenvironment{corollary}{
  enhanced jigsaw,
  boxrule=0pt, frame hidden,
  boxsep=0pt, top=4pt, bottom=4pt, left=8pt, right=8pt,
  arc=0pt, outer arc=0pt,
  colback=blue!4!white,
  borderline west={1.2pt}{0pt}{blue!50!cyan!60},
  before skip=8pt, after skip=8pt, breakable,
}
\tcolorboxenvironment{definition}{
  enhanced jigsaw,
  boxrule=0pt, frame hidden,
  boxsep=0pt, top=4pt, bottom=4pt, left=8pt, right=8pt,
  arc=0pt, outer arc=0pt,
  colback=blue!4!white,
  borderline west={1.2pt}{0pt}{blue!50!cyan!60},
  before skip=8pt, after skip=8pt, breakable,
}
\tcolorboxenvironment{proof}{
  enhanced jigsaw,
  boxrule=0pt, frame hidden,
  boxsep=0pt, top=4pt, bottom=4pt, left=8pt, right=8pt,
  arc=0pt, outer arc=0pt,
  colback=gray!6!white,
  borderline west={1.2pt}{0pt}{gray!55},
  before skip=8pt, after skip=8pt, breakable,
}
\newtcolorbox{leanbox}{
  enhanced jigsaw,
  colback=orange!4,
  boxrule=0pt, frame hidden,
  borderline west={1.2pt}{0pt}{orange!50!red!50},
  arc=0pt, outer arc=0pt,
  boxsep=0pt,
  top=4pt, bottom=4pt, left=8pt, right=6pt,
  before skip=4pt, after skip=4pt,
  breakable,
}

\title{Hypothesis-Disciplined Multi-Agent Automated Formalization of Asymptotic Statistical Theory}
\author{Tingzhou Wei$^{1}$, Zeyu Zheng$^{2}$, Ethan X. Fang$^{1}$, Junwei Lu$^{3}$\footnote{Lu is partially supported by NSF Artificial Intelligence, Formal Methods, and Mathematical Reasoning (AIMing) program DMS-2434664. Fang is partially supported by NSF grants DMS-2346292 and DMS-2434666.}\bigskip \\
\small 
$^1${Department of Biostatistics \& Bioinformatics, Duke University} \\
\small 
$^2${Department of Mathematical Sciences, Carnegie Mellon University} \\
\small
$^3${Department of Biostatistics, Harvard T.H. Chan School of Public Health}}
\date{}

\begin{document}
\maketitle

\begin{abstract}
Asymptotic statistical theory is a challenging domain for AI-assisted
formalization: its central results mix 
convergence statements, asymptotic expansions, functional analysis, and regularity conditions that have a large gap from existing infrastructure in Lean~4 formalization.
To address these challenges, we propose a
  hypothesis-disciplined Lean~4 formalization pipeline built from multiple agents:
a manager that coordinates seven specialist roles for proof planning,
skeleton scaffolding, Mathlib reconnaissance, proof construction,
integration, independent review, and audit.  The main methodological discipline is the
hypothesis-disciplined audit, implemented by the Auditor agent: every main-theorem hypothesis and
concept-layer field must be anchored in the source mathematical
prose, justified as a Lean encoding adapter, marked as source-implied,
or rejected as an unsupported strengthening.  Using this workflow, we
build a systematic formalization of
asymptotic statistical theory, especially the parametric and semi-parametric models' asymptotic distribution and efficiency results.  The resulting Lean development is axiom-clean and
source-faithful, with Lean-checked and human-audited proofs of core parametric and
semi-parametric theorems organized so that theorem-agnostic
infrastructure and statistical concept definitions are separated from
theorem-specific assembly. The formalization results are available in the following GitHub repository.
\begin{center}
\faGithub\ \textit{GitHub:}\ {\texttt{\href{https://github.com/junwei-lu/Lean-Asymptotic-Statistical-Theory}{https://github.com/junwei-lu/Lean-Asymptotic-Statistical-Theory}}}
\end{center}
\end{abstract}

\section{Introduction}
\label{sec:intro}

Asymptotic statistical theory provides the mathematical language for
large-sample inference: local experiments, likelihood expansions,
contiguity, weak limits of estimators, efficiency bounds, and
minimax risk comparisons \cite{vanderVaart1998,vanderVaart1996,
LeCam1986,Ibragimov1981}.  Formalizing this theory is therefore
not just a matter of proving a handful of named theorems; it
requires a reusable library of probability infrastructure and
statistical concepts such as parametric and semi-parametric
models, differentiability in quadratic mean, score functions,
Fisher information, kernels, and loss functionals.  Yet this
part of statistics has remained largely absent from
machine-checked mathematics.  Lean~4 and Mathlib
\cite{Moura2021Lean,Mathlib2020} provide substantial
measure-theoretic foundations and classical limit theorems, and
recent work has pushed agentic formalization at both the
textbook-chapter scale \cite{Gloeckle2026} and the level of
research-grade individual theorems \cite{Archon2026}, as well as
Lean-side statistical learning theory \cite{Zhang2026SLT}.  To our knowledge, there is no existing
automatic formalization pipeline for asymptotic statistical theory, especially the parametric and semi-parametric models' statistical efficiency.

There are several potential risks in the formalization of the asymptotic statistical theory.
Prior work formalizing non-asymptotic statistical learning theory \cite{Zhang2026SLT} observes in its Lean
formalization that statistical proofs written in natural language often leave
measurability and topological assumptions implicit and blur almost-sure
versus pointwise statements.  Asymptotic statistical formalization
inherits this ambiguity, and several of its structural features (e.g., ambiguity in asymptotic arguments, the regularity condition complexity, and the functional space issues)
sharpen it further.  In an agentic formalization pipeline, this ambiguity
turns into a concrete failure mode.  When an agent encounters a missing proof obligation it cannot
discharge, it can quietly absorb the obligation as an extra
hypothesis on the main theorem's signature or as an extra field on
a concept-layer definition.  The resulting Lean formalization may still
compile and be sorry-free.  We call the signature-level version of this
{hypothesis laundering} and the concept-level version
{definition drift}.  The latter is the former relocated to a definition:
the missing proof obligation is hidden as a concept-layer field or
helper definition rather than as a theorem hypothesis.  We
highlight four examples in the asymptotic statistical formalization below:
\begin{itemize}[leftmargin=*,labelindent=0pt,itemsep=2pt,topsep=4pt]
  \item \textbf{Natural-language imprecision at the formal boundary.}
    Source prose underspecifies choices Lean must commit to.
    ``The integral on the left'' may be pointwise, almost-sure,
    in-probability, or $L^p$; ``a measurable function'' may or
    may not require $L^2$-membership of its squares.  Each such
    phrase resolves to several distinct typed propositions and
    the agent must pick one.
  \item \textbf{Long Mathlib dependency chains amplify drift.}
    Statistical concepts stack deeply. For example, differentiability in quadratic mean
    sits on measurability, $L^p$-membership, and $\sigma$-finiteness, each stacking further down.  A single
    drifted field several layers up silently propagates to every
    downstream consumer.
  \item \textbf{Lean constructions without source anchor.}  Some
    constructions have no source-text syntactic counterpart at all:
    the asymptotic representation theorem requires realizing the
    limit law as the Gaussian-shift limit experiment composed with a
    Markov kernel; the prose gives only a randomized-statistic
    ($T(X,U)$) form, so the kernel object itself has nothing to
    transcribe directly.
  \item \textbf{Multi-source variation.}  The standard
    sources~\cite{vanderVaart1998,vanderVaart1996,LeCam1986,
    Ibragimov1981} each give the regularity condition with slightly different
    content, so a laundered hypothesis often reads to a reviewer as
    some other source's regularity: cross-wave drift
    across variants stays invisible without explicit version-tracking.
\end{itemize}

Our multi-agent formalization framework aims to make the source mathematical
prose the audit authority inside the multi-agent loop.  We therefore develop
an agent \textit{Auditor} for both locations of the same drift pattern, which
maintains a structured book-reference document for each audited theorem
boundary: every main-theorem hypothesis, instance constraint, and
concept-layer definition field is paired with a row classification, a
verbatim source excerpt, and a citation to the original section and
page in the chosen source (van~der~Vaart's {Asymptotic
Statistics}~\cite{vanderVaart1998} in our case).  We use
\textit{hypothesis-disciplined} in this broad boundary sense: the
audited obligations include theorem hypotheses, instance constraints,
and definition fields.  This row-level
artifact is not an after-the-fact checklist.  Immediately after the agent \textit{Scaffolder} fixes the opening skeleton, the agent \textit{Auditor} checks that the
main signature has source-backed rows; at each wave checkpoint, after
the agent \textit{Integrator} proposes a trunk state, it revisits both new rows and
historical rows affected by any changed signature or audited
definition; when an agent \textit{Executor} reports a genuine gap, the agent \textit{Auditor} first
classifies the proposed missing assumption before the agent \textit{Manager}
dispatches the next agent \textit{Planner} or agent \textit{Executor} task.  The result is a
hypothesis-disciplined workflow in which laundering is made visible at
the theorem boundary where it is introduced, and unsupported
strengthenings are removed, justified as Lean encoding adapters, or
escalated before they can propagate downstream.

On top of this discipline we develop a systematic Lean~4
formalization of central results in asymptotic statistical theory.  The
development covers parametric and semi-parametric models, with an
emphasis on asymptotic distributions, lower-bound phenomena, and
efficiency results for statistical procedures.  Its library is
organized into reusable theorem-agnostic infrastructure, statistical
concept definitions, and theorem-specific assemblies, so that later
formalization work in asymptotic statistics can reuse both the
mathematical components and the proof-engineering pattern.  At the
same time, the hypothesis-disciplined audit is text-agnostic by
construction: it requires only a chosen source text, row-level
evidence, and explicit classifications of formal obligations.  The
methodology therefore applies beyond the present case study to other
source-faithful formalization projects in deep mathematical domains
where important assumptions are implicit, layered, or easy to
strengthen inadvertently.

\subsection{Our Contributions}
\label{sec:contributions}

\noindent \textbf{A Lean 4 library of asymptotic statistical theory.}
The library contains a shared stratum of
Mathlib-side theorem-agnostic bricks
and a statistical concept stratum consisting of
asymptotic-statistics definitions together with reusable derived
properties of those definitions.  We present five cornerstone
theorems on both the parametric and semi-parametric sides; these
serve as named entry points for downstream formalization and are
not meant to delimit the library's reusable surface.

\noindent \textbf{A multi-agent automated formalization pipeline for thin-Mathlib-coverage infrastructure.}
Our second contribution is an execution scaffold for domains where the
formal library is too thin for a single proof-search loop to be the
right unit of work.  The architecture organizes the orchestration loop: one user-facing agent
\textit{Manager} assigns scoped tasks to specialist agents, isolates
their work in separate git worktrees, and admits progress only through
reviewed checkpoint merges on a buildable trunk.  This design turns
large formalization into a sequence of auditable engineering
transactions: reconnaissance can precede proof construction,
independent review can gate completed changes, and reusable
infrastructure can be promoted out of theorem-specific files as it is
discovered.  Our pipeline then provides a division of labor tailored to thin-Mathlib-coverage
formalization, where library search, proof decomposition,
cross-worktree deduplication, and shared-infrastructure growth must be
managed together.  

\noindent \textbf{An auditor to discipline against agent-side drift.}
We introduce a dedicated agent \textit{Auditor} that makes
source-faithfulness a runtime obligation rather than a post-hoc
judgment.  The \textit{Auditor} attaches source evidence to each theorem
boundary and audited definition boundary, so that a successful Lean
build is not accepted unless its assumptions are also justified by the
chosen mathematical source or by an explicit Lean encoding need.  It is
invoked exactly where agent-side drift tends to enter: after the
initial scaffold fixes a theorem statement, at wave checkpoints that
change audited boundaries, and when a proof obstruction pressures the
system to add a stronger assumption.  Unsupported assumptions are
therefore routed back to planning, repaired in the proof strategy, or
explicitly escalated rather than silently absorbed into the formal
statement. 

The rest of the paper is organized as follows.  Section~\ref{sec:arch}
describes the agent architecture and the source-reference
discipline.  Section~\ref{sec:domain} presents the Lean library and
the five highlighted cornerstones.  
Appendix~\ref{sec:evaluation} reports the scale and axiom
certificates and summarizes the hypothesis-audit outcomes, and
Appendix~\ref{app:case-studies} stress-tests the audit discipline
through two drift case studies.  Section~\ref{sec:conclusion}
concludes the paper and discusses the future work.

\subsection{Related Work}
\label{sec:related}

\noindent \textbf{Multi-agent Lean formalization.}
There are several existing works on the multi-agent Lean formalization, such as Gloeckle et
al.~\cite{Gloeckle2026}, who formalize an
algebraic combinatorics textbook with thousands of agents
collaborating on a shared code base via version control; Archon~\cite{Archon2026}, which orchestrates Claude
Code via a dispatcher targeting research-level individual
theorems; and Zhang et al.~\cite{Zhang2026SLT}, who formalize
non-asymptotic statistical learning theory under supervised Claude
Code use.  Our problem differs from each along the same axis: prior
systems emphasize scale, throughput, and proof closure; this
work adds a hypothesis-lineage audit, pairing mechanical
change-detection with a semantic faithfulness judgment, for source-faithful
textbook formalization in a thin-coverage, assumption-heavy
domain. We further use a multi-tool
Mathlib retrieval layer (type-by-name lookup, type-shape search, and
natural-language search) to match a higher-coverage setting.  There are also other recent works in this space: \cite{NuminaLeanAgent2026} is a general agentic reasoning system for formal mathematics, while \cite{hariharan2026milestone} and \cite{Ilin2026VML} focus on human-collaborative formalization of specific theorems rather than a systematic area of mathematical theory.

\noindent \textbf{Statement faithfulness and autoformalization drift.}
A parallel line studies a structurally different audit problem:
the gap between proof validity and translation faithfulness when a large language model
(LLM) produces a candidate Lean statement from natural language.
Kim et al.~\cite{KimPoirouxBosselut2026} document
formalization gaming; Li et
al.~\cite{Li2024SymbolicSemantic} score candidate formalization
by symbolic equivalence and back-translation similarity; Meadows et al.~\cite{MeadowsZhangFreitas2026} characterize 
semantic drift in physics auto-formalization.  Each operates at the
natural-language and Lean statement boundary on an
LLM-generated candidate.  Our novelty is a hypothesis-disciplined
audit for source faithfulness during collaborative theory
development: every theorem-boundary assumption and audited definition
boundary is tied to source evidence, a Lean encoding need, or an
explicit drift finding.  Faithfulness is therefore enforced as the
formal library evolves, rather than approximated by back-translation,
an LLM judge, or symbolic-equivalence scoring of a single candidate.
A separate ecosystem of benchmark datasets, tactic models, and
proof-search agents~\cite{Wang2025MALOT,Hilbert2025,AxProver2025,
ProverAgent2025,Apollo2025,MASA2025,LeanAgent2024,DeepSeekProverV2,
Goedel2025,LeanCopilot2024,LeanDojo2023} typically evaluates proof
search on single problems drawn from benchmarks. These systems operate on isolated theorems
with pre-supplied Mathlib-native formal statements and do not
address source-fidelity at theory-development scale.

\noindent \textbf{Formalized probability and statistics.}
In Lean, Sonoda et al.~\cite{Sonoda2025Rademacher} formalized
Rademacher complexity generalization bounds, followed by broader
non-asymptotic statistical learning theory in \cite{Zhang2026SLT}. Besides using Lean, there are other works formalizing probability and statistics in other interactive theorem provers. For example, in Rocq \cite{rocq_prover_9_2_0},
{Infotheo}~\cite{Infotheo} formalized discrete probability and
information theory and Affeldt et
al.~\cite{Affeldt2025Concentration} for non-asymptotic
concentration (Markov, Chebyshev, and Chernoff inequalities, culminating
in a Bernoulli-sampling bound).  In the {HOL4} theorem prover, some
probability theory topics are formalized \cite{HasanTahar2007ContPDF,
HasanTahar2007StdUniform}.  Using Isabelle/HOL \cite{nipkow2002isabellehol}, the
central limit theorem is formalized in \cite{AvigadHoelzlSerafin2017}, and the
L\'evy--Prokhorov metric and Prokhorov's
theorem are formalized in \cite{Hirata2024LevyProkhorov}.

\section{Multi-Agent Architecture}
\label{sec:arch}

\begin{figure*}[t]
\centering
\includegraphics[width=\textwidth]{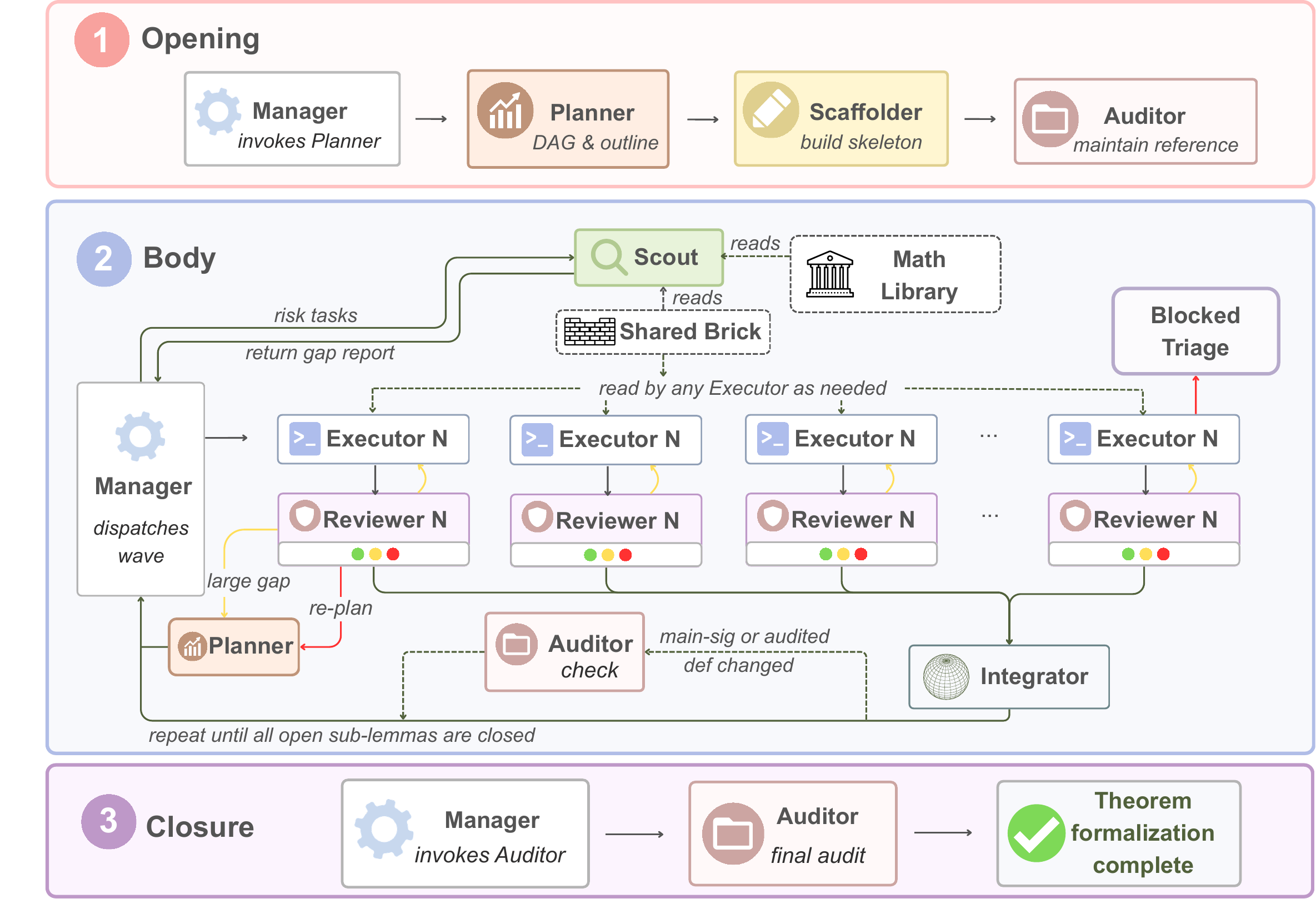}
\caption{Multi-agent architecture and runtime flow: a theorem's
  movement through the Opening, Body, and Closure phases, with the Body
  expanded into one parallel execution wave dispatched by the
  \textit{Manager} across the specialist roles.}
\label{fig:architecture}
\end{figure*}

This section describes the multi-agent architecture used to turn
source-faithfulness from a post hoc review concern into a runtime
discipline.  The system combines a fixed agent roster, an 
orchestration loop, and role-specific contracts for decomposing,
executing, reviewing, merging, and auditing proof work.  This design
responds to the two failure modes: hypothesis laundering at the main-theorem
signature and definition drift at the concept layer.  They share the
same source: definition drift is a proof obligation relocated from the
theorem signature into a project-authored concept definition, and is
therefore governed by the same hypothesis discipline.  Figure~\ref{fig:architecture} summarizes the architecture and the multi-agent orchestration flow.

The architecture of our formalization framework has one
user-facing parent agent and seven specialist agents that work
through task cards, isolated git worktrees, and disk-backed artifacts. We first introduce the agents as follows.
\begin{itemize}\itemsep1pt
  \item \textit{Manager}: It orchestrates the theorem workflow, dispatches
    all specialist agents, and decides the next step from their returned
    artifacts.
    \item \textit{Auditor}: It maintains book-reference rows for theorem
    boundaries and project-level definitions, and checks that signature
    or definition changes remain anchored in the source.
  \item \textit{Planner}: It converts the source proof into an executable
    sub-lemma plan and revises that plan when a route fails or needs
    finer decomposition.
  \item \textit{Scaffolder}: It emits the initial build-clean skeleton with
    sorry-stubbed bodies, thereby locking the sub-lemma
    signatures for later proof work.
  \item \textit{Scout}: It performs bounded Mathlib reconnaissance for
    risky sub-lemmas, records whether the needed bricks are present,
    composable, or absent, and truth-verifies suspect blocked goals.
  \item \textit{Executor}: It closes one assigned sub-lemma in an isolated
    worktree and returns a terminal state
    (see Table~\ref{tab:terminal-states}) with any candidate changes.
  \item \textit{Reviewer}: It gates each \textit{Executor} return by checking the
    candidate change set, build evidence, axioms usage output, and code
    quality before assigning a verdict.
  \item \textit{Integrator}: It folds accepted worktrees onto the trunk at
    wave checkpoints, rebuilds from scratch, deduplicates cross-worktree
    differences, and promotes reusable sub-lemmas into the shared library
    (Section~\ref{sec:domain}).
\end{itemize}

The detailed contracts for the four methodologically load-bearing roles
(\textit{Planner}, \textit{Scout}, \textit{Reviewer}, and
\textit{Auditor}) appear in Section~\ref{sec:agent-details}.

\begin{table}[t]
  \centering
  \small
  \setlength{\tabcolsep}{4pt}
  \newcolumntype{Y}{>{\raggedright\arraybackslash}X}%
  \begin{tabularx}{\linewidth}{@{}l l Y@{}}
    \toprule
    \textbf{State} & \textbf{Source} & \textbf{Actions} \\
    \midrule
    \multicolumn{3}{@{}l}{\textbf{Self-reported state on return}}\\
    \addlinespace[1.5pt]
    \texttt{DONE}        & \textit{Executor}          &
      Enters the review gate; the \textit{Reviewer} returns one of the three verdicts below \\
    \addlinespace[1.5pt]
    \texttt{PARTIAL}     & \textit{Executor}/\textit{Scout}  &
      Carries named residual sorries and enters the review gate \\
    \addlinespace[1.5pt]
    \texttt{BLOCKED}     & \textit{Executor}/\textit{Scout}  &
      No candidate changes; does not gate the checkpoint. Triaged by cause: a
      missing or suspect assumption goes to the \textit{Auditor}; a doubted
      goal is truth-verified by a \textit{Scout}, with counter-examples sent
      to the \textit{Planner} \\
    \addlinespace[1.5pt]
    \texttt{CLASSIFIED}  & \textit{Scout}             &
      Folds gap report into the revised plan \\
    \midrule
    \multicolumn{3}{@{}l}{\textbf{\textit{Reviewer} verdict on a candidate return}}\\
    \addlinespace[1.5pt]
    \texttt{GREEN}       & \textit{Reviewer}          &
      Mergeable: captured build green, body-only differences, baseline axioms only, no
      style or naming blocker \\
    \addlinespace[1.5pt]
    \texttt{YELLOW}      & \textit{Reviewer}          &
      Approach is sound but the proof is not yet fully closed: small residual
      sorries return to the same \textit{Executor} as a continuation, large ones to the
      \textit{Planner} for re-decomposition \\
    \addlinespace[1.5pt]
    \texttt{RED}         & \textit{Reviewer}          &
      Narrowing is redone by a fresh \textit{Executor} against the locked
      signature; a dead route goes to the \textit{Planner} for
      re-decomposition (rare safety net) \\
    \bottomrule
  \end{tabularx}
  \caption{Agent return states and the \textit{Manager}'s routing.  The returning agent
  self-identifies its state; a \texttt{DONE} or \texttt{PARTIAL} candidate
  additionally receives a \textit{Reviewer} verdict.}
  \label{tab:terminal-states}
  \vspace{-1em}
\end{table}

\subsection{Orchestration and Workflow}
\label{sec:orchestration}

A theorem target enters the system with its source material; at
opening time, the \textit{Manager} dispatches the \textit{Planner}, and the theorem
leaves as a build-clean Lean theorem.  Figure~\ref{fig:architecture}
breaks this movement into three lifecycle phases: \textbf{opening},
where the \textit{Planner} turns the source proof into an executable
sub-lemma plan; \textbf{body}, where the planned sub-lemmas are
closed; and \textbf{closure}, where the remaining proof debt and
book-reference audit are both discharged.  The body phase advances
by execution waves.  In one wave, the \textit{Manager} dispatches a
batch of \textit{Executor}s in parallel; each \textit{Executor}
closes one sub-lemma in its own worktree; a \textit{Reviewer} gates each
candidate return on the captured build and axioms usage
result; and the \textit{Integrator} folds the
accepted worktrees into one merge commit at the wave
checkpoint, building the merged result from scratch.  Green in
isolation does not imply green after merge, since cross-worktree
symbol collisions, signature drift, and the \textit{Integrator}'s own
deduplication and library promotion produce an artifact no \textit{Reviewer}
has built.  The trunk remains buildable at each checkpoint.

\noindent \textbf{Opening.} At opening time, the \textit{Manager} dispatches the \textit{Planner} on the
theorem target and source proof.  The \textit{Planner} drafts the informal
attack route---prose, an initial sub-lemma table, and a chosen
decomposition---and converts it into an executable plan: a
sub-lemma DAG with a per-row proof outline tying each formal
sub-lemma to the corresponding source-proof step.  The
\textit{Scaffolder} then commits a single
build-clean skeleton in which every sub-lemma body is
sorry, locking signatures for \textit{Executor} body-filling.  The
\textit{Auditor} performs its first book-reference pass on this
skeleton: it maintains the per-theorem reference rows for the
main-theorem signature and flags any assumption without a source
anchor.

\noindent \textbf{Body.} Waves run consecutively, each chosen from the open
sorries in the current scaffold.  In the baseline path, the \textit{Manager} selects a batch of sub-lemmas,
optionally sends the risky ones flagged by the \textit{Planner} to a
\textit{Scout} for Mathlib reconnaissance, and dispatches one
\textit{Executor} per sub-lemma.  When an \textit{Executor} returns a candidate
that the \textit{Reviewer} marks \texttt{GREEN}, it is accepted, and at
the wave checkpoint the \textit{Integrator} folds the wave's accepted
worktrees into one merge commit on the trunk; whenever that checkpoint
changes the main-theorem
signature or an audited definition, the \textit{Auditor} runs its
standing book-reference pass.  Two local
failures can pull a sub-lemma off this path: the \textit{Executor} cannot
produce a candidate, or the \textit{Reviewer} rejects the one it produced.  Both
route any source-fidelity question to the \textit{Auditor}, the project's main
safeguard against drift.  In rare cases a strategic re-plan resets the
route altogether.

\begin{figure}[t]
  \centering
  \includegraphics[width=0.65\textwidth]{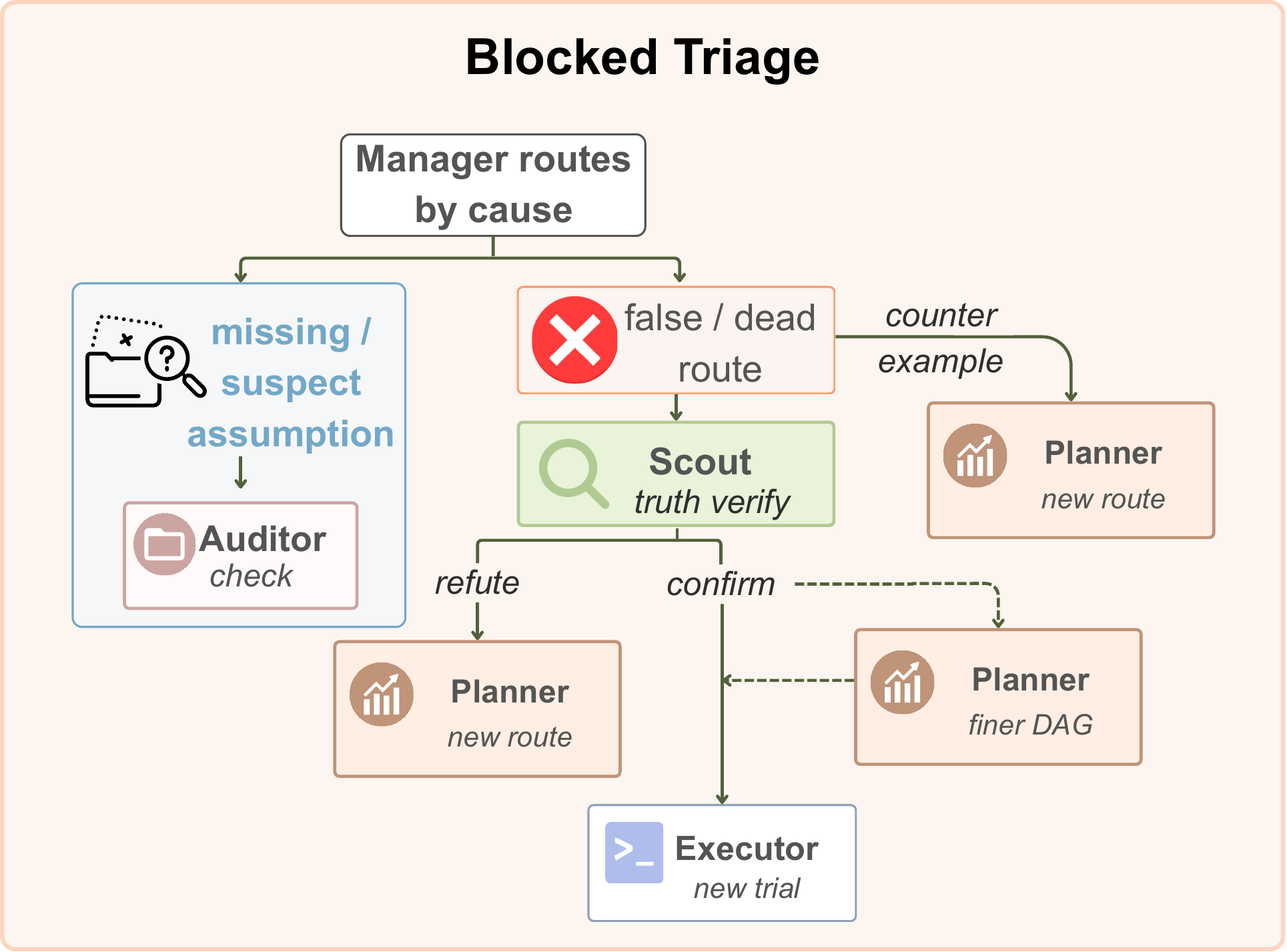}
  \caption{Triage of a blocked \textit{Executor} return: a missing or
    suspect assumption routes to the \textit{Auditor}'s drift control,
    while a doubted sub-lemma is truth-verified by a \textit{Scout}.}
  \label{fig:blocked}
\end{figure}
\noindent $\bullet$ \underline{\textit{Executor} block.}
An \textit{Executor} that cannot produce a candidate at all returns
\texttt{BLOCKED} with a diagnosis and evidence; Figure~\ref{fig:blocked}
summarizes how the \textit{Manager} then routes the blocked sub-lemma by
cause.  It carries no
candidate changes, so it bypasses the \textit{Reviewer} and does not block the
checkpoint: the \textit{Integrator} still folds the wave's other accepted
worktrees, while the \textit{Manager} triages the blocked sub-lemma for
reassignment or replanning.  A missing or suspect
assumption, a hypothesis the proof needs but the signature lacks, goes
to the \textit{Auditor}'s drift control.  When the \textit{Executor} instead doubts the
sub-lemma itself (it may be false, or the route may not reach it), a
\textit{Scout} truth-verifies the goal: a refutation, or a counter-example the
\textit{Executor} already holds, returns it to the \textit{Planner} for a new route,
while a confirmation hands a fresh \textit{Executor} the supporting brick, or,
when the confirmed goal still needs breaking down, prompts the \textit{Planner}
to re-decompose it into a finer DAG.

\noindent $\bullet$ \underline{\textit{Reviewer} rejection.}
After review, a candidate return (\texttt{DONE} or \texttt{PARTIAL})
that is not \texttt{GREEN} receives one of two non-merging verdicts.  \texttt{YELLOW}
means that the route is sound but named residual sorries remain; this
is the normal verdict for a \texttt{PARTIAL} return unless review finds
a narrowing or dead route.  Small residuals then return
to the same \textit{Executor}, large ones go to the \textit{Planner} for
re-decomposition, and the closed part still merges.  \texttt{RED} marks a
candidate that proves a narrower or different statement than the locked
signature, or whose route is dead; the \textit{Reviewer} contract
(Section~\ref{sec:reviewer}) specifies how each is routed.

\noindent $\bullet$ \underline{Strategic re-plan.}
Occasionally local repairs can no longer preserve the chosen route.  In
that case a strategic re-plan, authorized by the user, makes a
wholesale change of proof strategy.  Triggered by a falsification, a
scope correction, or accumulated drift findings, it resets the
sub-lemma DAG and emits a fresh \textit{Scaffolder} skeleton, and unlike the
within-strategy repairs above it crosses back to the user.  

\textit{Planner} reruns can interleave with new wave dispatches rather than
blocking them, so a re-decomposition for one sub-lemma proceeds while
others advance.  These repairs all stay within the body phase of an
already-open theorem, in contrast to the strategic re-plan above.

\noindent \textbf{Closure.}
Closure is the state in which the sorry count on the main
theorem's transitive closure reaches zero, and the \textit{Auditor}'s standing
seven-check passes (Section~\ref{sec:auditor}); the same audit that
accompanies wave-boundary merges certifies closure when the last
sorry disappears.  The loop provides three structural
guarantees: accepted work reaches the trunk only at wave checkpoints,
where it stays buildable; the merged trunk is rebuilt from scratch by
the \textit{Integrator} at each checkpoint, while the \textit{Reviewer} confirms every
candidate's axioms usage output against the accepted
baseline; and every committed change
to a main-theorem signature or audited definition is checked against
the \textit{Auditor}'s reference rows.

\subsection{Agent Contracts}
\label{sec:agent-details}

The four agents whose contracts carry the load on the
methodological discipline are spelled out in turn.  The remaining
roles (\textit{Scaffolder}, \textit{Executor}, \textit{Integrator}, \textit{Manager}) operate by the
lifecycle described in Section~\ref{sec:orchestration} and need no
contract beyond their inventory entry.

\subsubsection{Auditor Agent}
\label{sec:auditor}
\label{sec:hypothesis-discipline}
\label{sec:audit}

\begin{figure}[t]
  \centering
  \includegraphics[width=0.75\textwidth]{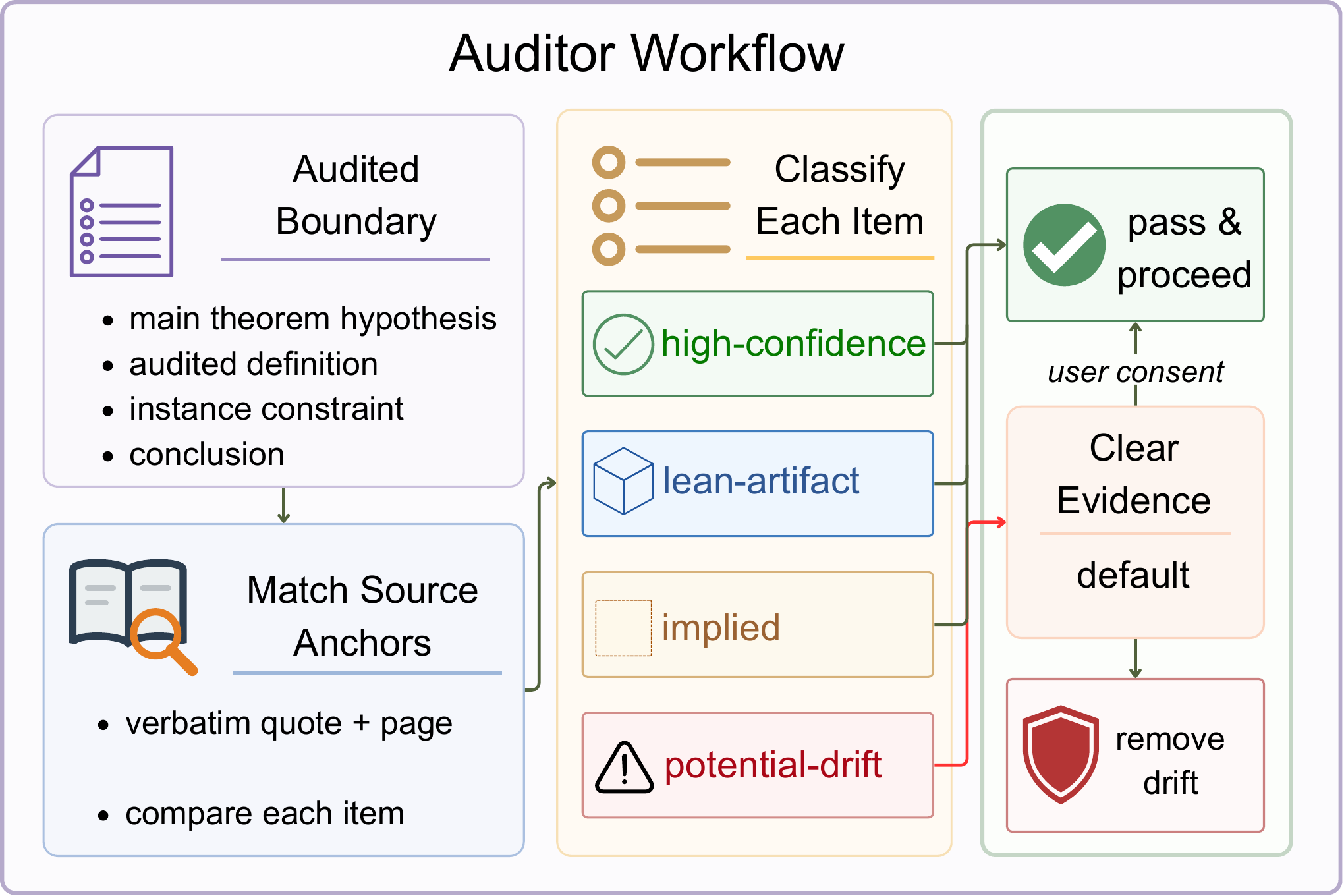}
  \caption{The \textit{Hypothesis-Disciplined Auditor}'s source-anchor classification: each
    audited main-theorem hypothesis, instance constraint, and definition field is tied to a
    book-reference row, classified by its source anchor, and either
    accepted (directly or by user consent) or removed as drift.}
  \label{fig:auditor-workflow}
\end{figure}

The \textit{Auditor} maintains the book-reference rows for main-theorem
hypotheses, instance constraints, and project-level definitions.  For each audited Lean
entry point it records the row classification, source citation, and
evidence, then checks whether each audited entry has been
strengthened beyond its source anchor.  Its output is the
row-level finding used by the \textit{Manager} at wave boundaries.
Figure~\ref{fig:auditor-workflow} summarizes this source-anchor
classification, in which each audited row is either accepted (directly,
or by user consent when the source anchor is not clear) or removed as
drift.

The canonical audit, fired every time the main-theorem signature or
an audited definition changes, is seven checks.  Each check separates a
mechanical detection step, which surfaces that something changed,
from a semantic judgment step, in which the \textit{Auditor} agent
decides whether the surfaced change is faithful to the source, removing
an unanchored assumption by default and reaching the user only when
evidence shows it is genuinely needed.  Two checks are
fully mechanical (1--2); four pair mechanical change-detection with a
per-item semantic faithfulness judgment (3--6); one is a global semantic
spot check (7).  The
mechanical detection under checks~3--6 is a comparison against the
committed signature-pinning snapshot (Appendix~\ref{app:signature-pinning}),
which surfaces any change to
a hypothesis, the conclusion, an instance binder, or an authored
definition body.  See Table~\ref{tab:audit-checks} for details.

\begin{table}[t]
  \centering
  \small
  \renewcommand{\arraystretch}{1.15}
  \begin{tabular}{@{}c l l l@{}}
    \toprule
    \# & Check & Kind & Trigger \\
    \midrule
    1 & no stray sorry & mechanical & per build \\
    2 & axioms usage clean & mechanical & per build \\
    3 & each main-sig hypothesis matches a reference row & semantic, per-item & main-sig change \\
    4 & conclusion equivalent to source & semantic, per-item & main-sig change \\
    5 & each instance constraint matches a row & semantic, per-item & main-sig change \\
    6 & \texttt{structure} fields, \texttt{def} edges classified & semantic, per-def & def change \\
    7 & main assumptions jointly satisfiable & semantic, global & hyp.\ revision \\
    \bottomrule
  \end{tabular}
  \caption{Seven checks performed by the hypothesis-disciplined audit, including
    explicit accounting for instance constraints.
    Mechanical checks are decided by tooling; semantic checks are the
    \textit{Auditor} agent's source-faithfulness calls.}
  \label{tab:audit-checks}
  \vspace{-1em}
\end{table}

At a wave checkpoint, the \textit{Manager} schedules this pass after the
\textit{Integrator} has produced a candidate trunk state whose audited boundary
changed: a new or rewritten main-theorem signature, or a modified
project-authored definition on the audited dependency path.  The pass
revisits both rows introduced by the current wave and historical rows
that may have gone stale after a refactor.  The audit tool below
supplies the mechanical side of these checks.

\noindent \textbf{Lifecycle role.}
The \textit{Auditor} is invoked at the points where the theorem boundary can
drift.  Immediately after the \textit{Scaffolder} commits the opening skeleton,
it runs the book-reference pass: it creates or updates the reference
rows for the main-theorem signature, checks that every row has a
verbatim quote and page citation, and pushes the \textit{Planner} to eliminate
any \texttt{potential-drift} assumption carried over from the informal
outline.
At wave checkpoints it performs the pass described above.  When a
\texttt{BLOCKED} report raises a missing or suspect assumption, it
routes through the \textit{Auditor} first, which classifies the assumption
before the \textit{Manager} chooses the next \textit{Planner} or \textit{Executor} dispatch.

\noindent \textbf{Hypothesis taxonomy.}
Every signature argument lands in one of four tiers; the
book-reference row records the tier.

\begin{itemize}\itemsep1pt
  \item \texttt{high-confidence}: appears verbatim in the
    book (header or proof body).  Row must cite the book's
    chapter, page, and verbatim quote.
  \item \texttt{lean-artifact}: encoding adapter required by
    Lean / Mathlib API (typeclass instance, joint-measurability
    witness, $L^2$-membership for Bochner integrability, support
    bookkeeping for change-of-measure, and similar).  Each row
    must cite a {concrete} encoding correspondence (e.g.\
    ``the \texttt{Kernel.compProd} construction requires joint
    measurability of $(\theta, x) \mapsto p_\theta(x)$''); the
    phrase ``implicit in book'' is forbidden as the sole
    justification.
  \item \texttt{implied}: the book sentence implies the
    hypothesis without stating it (e.g.\ ``has covariance
    matrix'' semantically presupposes a finite second moment).
    Auto-accepted on a citation to the implying sentence.
  \item \texttt{potential-drift}: no book anchor.
    The \textit{Manager} first treats the row as a route error and asks the
    \textit{Planner} or \textit{Executor} to remove the assumption or repair the
    statement.  If a counter-example, or a sustained proof obstruction
    diagnosed by the agents, indicates that the assumption is needed
    for the proposed theorem, the row escalates to the user; accepted
    rows stay \texttt{potential-drift} with the evidence recorded and
    explicit sign-off.
\end{itemize}

\noindent \textbf{Definition discipline.}
The same machinery applies to project-authored \texttt{structure}
and \texttt{def} declarations on the audited dependency path.  A
definition-side row is judged by the same source-anchor test as a
signature row: the field or definition edge must appear in the source
definition, be implied by it, or be justified as a Lean encoding
adapter.  This is not a parallel discipline: it is the place where a
hypothesis-like obligation can be relocated into a definition.  If it
has no such anchor, it is definition-side \texttt{potential-drift}.  The
typical drift pattern is relocation: an extra condition used by a proof route is encoded as a
\texttt{structure} field or a project-authored helper
\texttt{def}.  The \textit{Auditor} records such rows in the same
book-reference document and classifies them by the source anchor that
justifies the field or definition.

\noindent \textbf{Audit tool.}
A lightweight audit-tool suite complements the \textit{Auditor}'s per-item semantic checks.
Its passes mirror the mechanical side of the seven-check audit:
a build-hygiene pass counts stray \texttt{sorry} occurrences and reports
their transitive taint; a signature-pinning pass compares the audited
theorem's fully elaborated signature, together with the bodies of the
authored definitions it depends on, against a committed snapshot, so any
change to a hypothesis, an instance argument, the conclusion, or a
definition body is surfaced
(Appendix~\ref{app:signature-pinning}); and a declaration-inventory pass
enumerates project-authored \texttt{structure}s, \texttt{def}s,
and related declarations in the audited scope.  Against the book-reference document, the tool runs a coverage check on
the signature's hypotheses: it flags any hypothesis with no matching
row, any stale row, the forbidden ``implicit'' / ``by analogy'' /
``book is informal'' defenses on \texttt{high-confidence} rows, and any
\texttt{potential-drift} row still present among the theorem's formal
assumptions.  This matches hypothesis names against the document's
rows, not against \cite{vanderVaart1998}.  A \texttt{potential-drift} row accepted by the user
remains in the reference document with its evidence and sign-off, so
later audits can distinguish it from unresolved drift.
The \textit{Auditor} agent makes the source-faithfulness, row
classification, and joint-consistency judgments, reading the candidate
items and reproducible checks that the tool produces to reach its
verdict.  In our development this resolved most discrepancies at the
agent level, without escalation to the user.

\subsubsection{Planner Agent}
\label{sec:planner}

The \textit{Planner} owns proof planning.  At opening time, it reads the
theorem target and source proof, drafts the attack route, and turns
that route into a sub-lemma DAG with a per-row proof outline.  When
the source proof invokes a supporting result or proof device, the
\textit{Planner} records it either as an available brick or as a local
derivation target on the DAG.

The same role handles planning repairs once the body phase is under
way.  A \texttt{YELLOW} verdict with large residual sorries, or
a \texttt{BLOCKED} goal that a \textit{Scout} verifies as true but reachable
only through a finer breakdown, asks the \textit{Planner} to refine the
remaining lemma into a finer DAG while preserving the existing
\texttt{GREEN} sub-lemmas.  A \texttt{RED} verdict, or a \texttt{BLOCKED}
route falsified by a counter-example or a \textit{Scout}'s truth-verification,
asks it instead to abandon the current route and draft a new one.

\subsubsection{Scout Agent}
\label{sec:scout}

A \textit{Scout} is a bounded reconnaissance probe: given one sub-lemma the
\textit{Planner} has flagged as risky, it spends a fixed time budget
checking what Mathlib actually supplies, and returns a gap report
with cited evidence.  The role exists because thin-coverage domains
make coverage uncertainty load-bearing: whether the bricks a
sub-lemma needs are present, composable from existing pieces, or
absent entirely.

During \texttt{BLOCKED} triage, a \textit{Scout} also truth-verifies a
suspect goal under the locked signature: it either confirms the goal
and names the supporting brick or reports a refutation or counter-example
that sends the route back to the \textit{Planner}.

\noindent \textbf{Risk classification.}
The \textit{Planner} marks a sub-lemma risky when its closure depends on
Mathlib API of uncertain existence or shape: results named in the
source proof (central limit theorems, Prokhorov, Le~Cam lemmas)
whose informal name may not correspond to a discoverable Mathlib
declaration, theorem-sized targets that are unlikely to be packaged
under one declaration, and steps where the right brick exists but
its hypotheses may not match the sub-lemma's assumptions.
Sub-lemmas with no such dependency skip reconnaissance and go
straight to an \textit{Executor}.

\noindent \textbf{Gap report.}
A \textit{Scout}'s success state is \texttt{CLASSIFIED}: for each expected
brick it records whether the target is directly supported
(the same objects, operation, and conclusion already in Mathlib),
composable (the ingredients exist and the remaining gap is
routine adapter work), or not found (no declaration and no
sufficient ingredients), with the evidence cited.  The report also
flags adapter gaps (a brick whose hypotheses need bridging),
critiques the scaffold's locked signature when it does not express
the intended mathematics, estimates the cost to close, and names
fragments that should be promoted to the shared library.

When a gap is caught before \textit{Executor} work begins, the
\texttt{CLASSIFIED} report can redirect the \textit{Planner}; the same gap
found inside an \textit{Executor} attempt usually forces a downstream
re-plan.  The report is recorded for reuse by later \textit{Scout}s and
\textit{Executor}s, but route selection stays with the \textit{Planner}.

\subsubsection{Reviewer Agent}
\label{sec:reviewer}

The \textit{Reviewer} gates each candidate return (\texttt{DONE} or
\texttt{PARTIAL}) and issues the verdict that routes it.  Its work is
to compare the candidate change set against the locked scaffold, review the code for
style, and write a pull request-style change summary.  The change set is what certifies
that the proof closes the assigned statement rather than a weaker one:
the \textit{Scaffolder}'s skeleton pins each sub-lemma signature, a well-formed
candidate changes only the body, and a body-only change that typechecks
proves exactly the locked statement, since Lean admits no narrower
proof against a fixed signature and fixed referenced definitions.  Any
narrowing therefore shows up in the change set as an edited signature
line, or---when relocated into a project-authored definition the
signature references---as an edited audited definition (the
definition-drift channel of Section~\ref{sec:auditor}).  The candidate's axioms usage
output surfaces any residual sorry or non-baseline axiom.  The
verdict is one of three:

\begin{itemize}\itemsep1pt
  \item \texttt{GREEN} means the change is mergeable: the captured
    \texttt{lake build} is green, the change touches only bodies, axioms usage on the
    target contains only the baseline axioms, and the code review
    finds no style or naming blocker.
  \item \texttt{YELLOW} means the approach is sound but the proof is
    not yet closed: named residual sorries remain, so small
    residuals return to the same \textit{Executor} for continuation and larger
    ones go to the \textit{Planner} for re-decomposition.  A \texttt{PARTIAL}
    return lands here by construction.
  \item \texttt{RED} means the candidate proves a narrower or different
    statement than the locked signature, or that its route is dead.  The
    \textit{Manager} either reassigns the work to a fresh
    \textit{Executor} against the still-standing signature or sends it
    to the \textit{Planner} for re-decomposition, escalating to a
    strategic re-plan when local repair no longer preserves the route.
    \texttt{RED} is a rare safety net for a
    confident-but-wrong submission: most route failures surface upstream
    as \texttt{BLOCKED}, and the locked-signature discipline pre-empts
    most narrowing.  It did fire in our development, catching a sub-lemma
    whose signature had drifted from its locked form.
\end{itemize}

The book-reference row check is conditional.  Ordinary sub-lemma body
changes do not consult the book-reference document; when a candidate
changes the main-theorem signature or an audited definition, the
\textit{Reviewer} compares only the changed main-signature hypotheses,
definition fields, and definition edges against the per-theorem
book-reference document (Section~\ref{sec:auditor}), and a
main-signature hypothesis or definition-side field lacking a matching
row blocks the merge until the \textit{Auditor} creates or updates the
corresponding row.

\section{Formalization Results for Asymptotic Statistical Theory}
\label{sec:domain}

In this section, we use five cornerstone theorems and several core
definitions from the resulting Lean library to make the formalization
artifact concrete.  These examples show how asymptotic statistics
theory is represented in Lean through reusable statistical concept
definitions and theorem-level Lean signatures.  They also identify
recurring formalization challenges: implicit source-proof assumptions,
Lean representation choices, proof-body hypotheses promoted to typed
inputs, and potential drift between the source argument and the formal
statement. Throughout this section we follow van~der~Vaart's \emph{Asymptotic
Statistics}~\cite{vanderVaart1998} as the source text for the theorems and definitions we formalize.

\begin{figure}[!htbp]
\centering
\includegraphics[width=\textwidth]{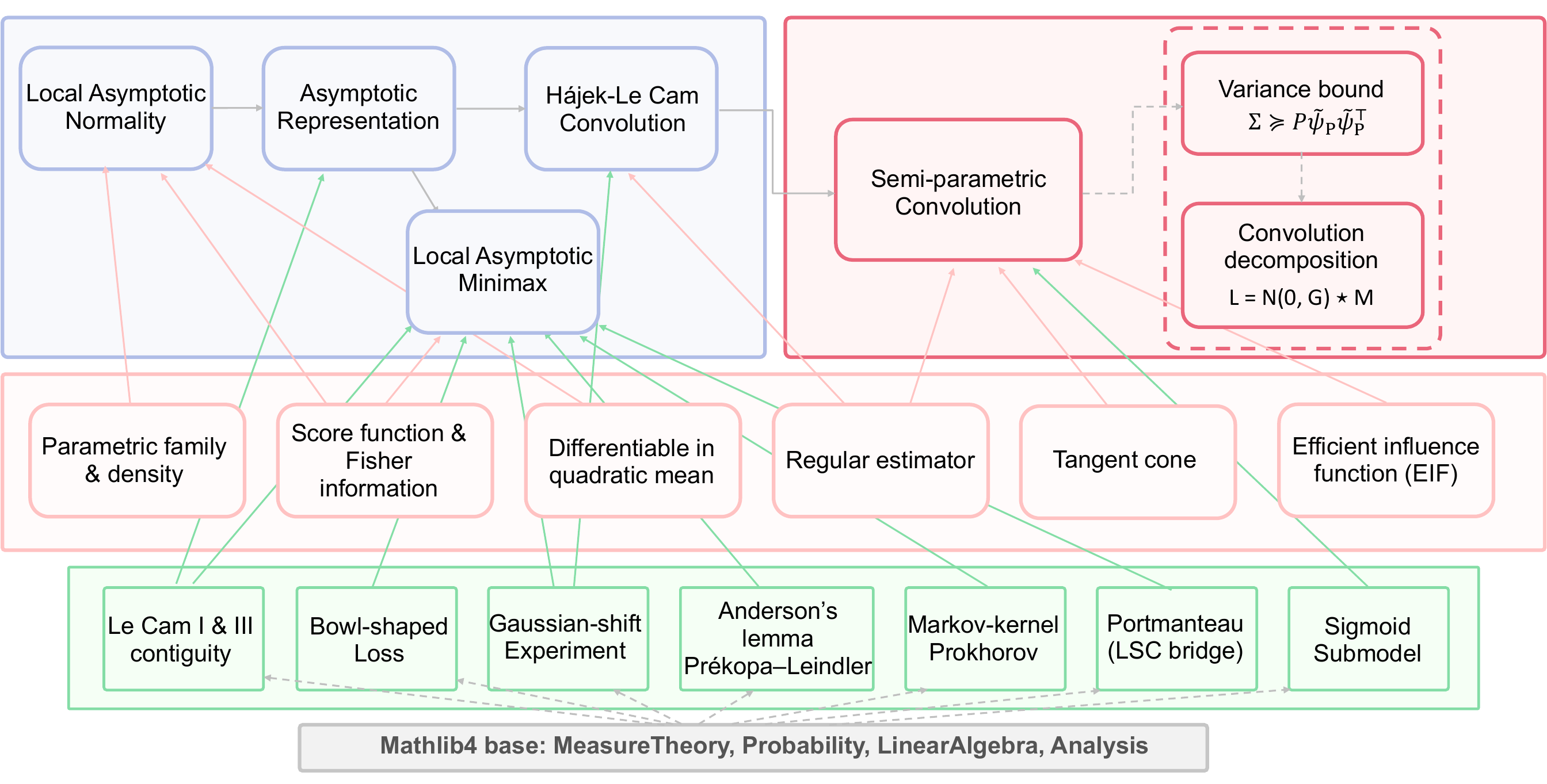}
\caption{Dependency graph of the five cornerstone theorems with
approximate per-cornerstone line counts.  The upper-left blue panel
groups the four parametric cornerstones; the upper-right rose panel
shows the semi-parametric cornerstones.  The
middle band lists statistical concept-layer definitions, the
lower green band lists reusable shared-library bricks, and the gray
baseline records the Mathlib4 dependencies.}
\label{fig:theorem-dag}
\end{figure}

\subsection{The Statistical Concept Layer}
\label{sec:concept-layer}

Before turning to individual cornerstones, we describe the reusable
library objects that those cornerstones share.  Mathlib4 supplies
measure theory and the classical CLT but not the parametric-family
machinery for asymptotic estimation, so the released library separates
the reusable material by import direction.  The shared-library stratum
contains theorem-agnostic analytic, measure-theoretic, and probability
infrastructure absent from current Mathlib.  The statistical concept
layer builds on this infrastructure to define the asymptotic-statistics
objects whose fields encode model, estimator, and tangent-space
definitions.  The separation follows the Lean dependency role rather
than whether a predicate appears in a theorem statement.  The
cornerstone demonstrations then instantiate these reusable components
in five released entry-point theorem files, while the library also
contains many supporting theorems and reusable intermediate results.

Figure~\ref{fig:theorem-dag} shows the reusable objects that enter the
cornerstone signatures, split by this dependency direction into
concept-layer definitions and shared-library bricks.  We expand the
three concept-layer definitions that most directly explain the surfaced
assumptions below:
Differentiable Quadratic Mean, regular estimator sequence, and tangent
set with its generated tangent space.

\vspace{-1em}
\paragraph{$\bullet$ Differentiability in quadratic mean.}
We use the quadratic-mean differentiability condition of Chapter~7, Eq.~(7.1) of \cite{vanderVaart1998}.
\begin{definition}
A model with densities $p_\theta$ is differentiable in quadratic
mean at $\theta$ if there exists a measurable score vector
$\ell_\theta$ such that
\[
  \int \left(
      \sqrt{p_{\theta+h}(x)} - \sqrt{p_\theta(x)}
      - \tfrac{1}{2}\langle h, \ell_\theta(x)\rangle
        \sqrt{p_\theta(x)}
  \right)^2\,d\mu(x)
  = o(\|h\|^2)
  \qquad (h \to 0).
\]
\end{definition}

Lean represents the chosen score vector as an explicit parameter
\(\ell\) of the structure.  The structure then has two fields:
\verb|isLittleO| records the displayed rate statement, and \verb|mem|
records that the residual is eventually an $L^2(\mu)$ function.
\begin{leanbox}
\small
\begin{verbatim}
structure DifferentiableQuadraticMean
    (M : ParametricFamily 𝓧 Θ) (μ : Measure 𝓧) (θ : Θ)
    (ℓ : 𝓧 → Θ) : Prop where
  mem : ∀ᶠ h in 𝓝 (0 : Θ),
    MemLp (fun x => M.sqrtDensity (θ + h) x - M.sqrtDensity θ x
                    - (1/2 : ℝ) * ⟪h, ℓ x⟫ * M.sqrtDensity θ x) 2 μ
  isLittleO :
    (fun h : Θ =>
      ∫ x, (M.sqrtDensity (θ + h) x
            - M.sqrtDensity θ x
            - (1/2 : ℝ) * ⟪h, ℓ x⟫ * M.sqrtDensity θ x) ^ 2 ∂μ)
    =o[𝓝 (0 : Θ)] (fun h : Θ => ‖h‖ ^ 2)
\end{verbatim}
\end{leanbox}
\noindent The Lean definition uses Mathlib's \(L^2\)-membership
predicate and asymptotic little-\(o\) notation.  The field
\verb|isLittleO| records the displayed rate statement.  The field
\verb|mem| records the integrability content presupposed by the source
integral: Mathlib's Bochner integral is defined even for
non-integrable functions, so the rate statement alone would not force
the residual to be an \(L^2\) object.

\vspace{-1em}
\paragraph{$\bullet$ Regular estimator sequence.}
We use the regular-estimator definition of Chapter~8 of \cite{vanderVaart1998}.
\begin{definition}
A sequence of statistics $T_n$ is \emph{regular} at $\theta_0$ for
estimating $\psi(\theta_0)$ if there exists a probability measure
$L_{\theta_0}$ such that, for every $h \in \mathbb R^k$,
\[
  \sqrt n\bigl(T_n - \psi(\theta_0 + h/\sqrt n)\bigr)
  \rightsquigarrow L_{\theta_0}
  \quad\text{under } P^n_{\theta_0 + h/\sqrt n}.
\]
The same limit law $L_{\theta_0}$ is used for every $h$.
\end{definition}

Lean represents this source condition as the following structure.
\begin{leanbox}
\small
\begin{verbatim}
structure RegularEstimatorSequence
    (M : ParametricFamily 𝓧 (AsymptoticRepresentation.Θ k))
    (μ : Measure 𝓧)
    (θ₀ : AsymptoticRepresentation.Θ k)
    (ψ : AsymptoticRepresentation.Θ k → AsymptoticRepresentation.𝓨 d)
    (T : ∀ n, (Fin n → 𝓧) → AsymptoticRepresentation.𝓨 d) :
    Type where
  limitDist : Measure (AsymptoticRepresentation.𝓨 d)
  isProb : IsProbabilityMeasure limitDist
  tendsto : ∀ h : AsymptoticRepresentation.Θ k,
    WeakConverges
      (fun n : ℕ =>
        (AsymptoticRepresentation.productMeasure M μ
            (θ₀ + (Real.sqrt n)⁻¹ • h) n).map
          (fun x => (Real.sqrt n) •
            (T n x - ψ (θ₀ + (Real.sqrt n)⁻¹ • h))))
      limitDist
\end{verbatim}
\end{leanbox}
\noindent The Lean definition uses Mathlib measures together with the
project's \verb|WeakConverges| predicate.  The fields
\verb|limitDist| and \verb|tendsto| record the common weak-limit law
and the convergence required for every local perturbation \(h\);
\verb|isProb| records that this limit law is a probability measure.

\vspace{-1em}
\paragraph{$\bullet$ Tangent set and tangent space.}
We use the tangent-set discussion and closed-linear-span construction of Chapter~25.3 of \cite{vanderVaart1998}.
\begin{definition}
At $P$, a \emph{tangent set} \(\dot{\mathcal P}_P\) is a collection
of score functions obtained from quadratic-mean-differentiable
submodels through $P$; by Lemma~25.14, these scores lie in
$L^2_0(P)$.  The \emph{generated tangent space} is the closed linear
span
$
  \operatorname{lin}\dot{\mathcal P}_P
$
of this tangent set.
\end{definition}

The source first names the tangent set and later uses its closed
linear span to define efficient influence functions and lower-bound
statements.  Lean keeps these two source-level objects separate:
\verb|TangentSpec| formalizes \(\dot{\mathcal P}_P\), recording the
tangent directions together with a realizing submodel for each
direction, while \verb|tangentSpace| formalizes
\(\operatorname{lin}\dot{\mathcal P}_P\).
\begin{leanbox}
\small
\begin{verbatim}
structure TangentSpec (P : Measure Ω) [IsProbabilityMeasure P] where
  carrier : Set ↥(L2ZeroMean P)
  submodelOf : ∀ g ∈ carrier,
    ∃ γ : AsymptoticStatistics.Core.QMDPath.QMDPath P, γ.score = g

noncomputable def tangentSpace
    {P : Measure Ω} [IsProbabilityMeasure P] (T : TangentSpec P) :
    Submodule ℝ ↥(L2ZeroMean P) :=
  (Submodule.span ℝ T.carrier).topologicalClosure
\end{verbatim}
\end{leanbox}
\noindent The Lean definition uses Mathlib's \(L^2\) space and
closed-submodule operations.  In \verb|TangentSpec|, \verb|carrier| is
the tangent set itself, while \verb|submodelOf| records that each listed
direction is realized as the score of a quadratic-mean-differentiable
submodel.  The derived object \verb|tangentSpace| then takes the closed
linear span of \verb|carrier|, corresponding to
\(\operatorname{lin}\dot{\mathcal P}_P\) in the source.

\subsection{Local Asymptotic Normality}
\label{sec:lan}

Local Asymptotic Normality (LAN) expresses the log-likelihood ratio
of a regular parametric experiment under local reparametrization
as a quadratic-Gaussian-plus-remainder; it is the gateway through
which every downstream cornerstone reduces to a Gaussian-shift
question.

\begin{theorem}[{\cite[Theorem~7.2]{vanderVaart1998}}]
Suppose $\Theta$ is an open subset of $\mathbb R^k$ and the model
$(P_\theta : \theta \in \Theta)$ is differentiable in quadratic
mean at $\theta$.  Then $P_\theta \ell_\theta = 0$ and the Fisher
information matrix $I_\theta = P_\theta \ell_\theta
\ell_\theta^\top$ exists.  Furthermore, for every converging
sequence $h_n \to h$, as $n \to \infty$,
\[
  \log \prod_{i=1}^n \frac{p_{\theta + h_n/\sqrt n}}{p_\theta}(X_i)
    = h^\top \Delta_{n,\theta}
       - \tfrac12 h^\top I_\theta h
       + o_{P_\theta}(1),
\]
where $\Delta_{n,\theta} = n^{-1/2} \sum_{i=1}^n \ell_\theta(X_i)$
is asymptotically $N(0, I_\theta)$.
\end{theorem}

The Lean entry point is \texttt{LAN\_expansion}, with signature:
\begin{leanbox}
\small
\begin{verbatim}
theorem LAN_expansion
    {k : ℕ}
    {Ω : Type*} {mΩ : MeasurableSpace Ω} (P : Measure Ω)
    [IsProbabilityMeasure P]
    (M : ParametricFamily 𝓧 (EuclideanSpace ℝ (Fin k))) (μ : Measure 𝓧)
    [SigmaFinite μ]
    (θ₀ : EuclideanSpace ℝ (Fin k))
    (ℓ : 𝓧 → EuclideanSpace ℝ (Fin k)) (hℓ : Measurable ℓ)
    (hPDF : IsPDFOf M μ)
    (hDQM : DifferentiableQuadraticMean M μ θ₀ ℓ)
    (h : EuclideanSpace ℝ (Fin k)) (h_n : ℕ → EuclideanSpace ℝ (Fin k))
    (hconv : Filter.Tendsto h_n Filter.atTop (𝓝 h))
    (X : ℕ → Ω → 𝓧) (hX_meas : ∀ i, Measurable (X i))
    (hindep : Pairwise fun i j => ProbabilityTheory.IndepFun (X i) (X j) P)
    (hident : ∀ i, ProbabilityTheory.IdentDistrib (X i) (X 0) P P)
    (hlaw : Measure.map (X 0) P
              = μ.withDensity fun x => ENNReal.ofReal (M.density θ₀ x)) :
    (∀ u, ∫ x, ⟪u, ℓ x⟫ * M.density θ₀ x ∂μ = 0) ∧
    (∀ u, Integrable (fun x => ⟪u, ℓ x⟫^2 * M.density θ₀ x) μ) ∧
    TendstoInMeasure P
      (fun n ω =>
        (∑ i ∈ Finset.range n,
          Real.log (M.density (θ₀ + (Real.sqrt n)⁻¹ • h_n n) (X i ω)
                    / M.density θ₀ (X i ω)))
        - (Real.sqrt n)⁻¹ * ∑ i ∈ Finset.range n, ⟪h, ℓ (X i ω)⟫
        + (1/2 : ℝ) * fisherInformation M μ θ₀ ℓ h h)
      Filter.atTop (fun _ => (0 : ℝ))
\end{verbatim}
\end{leanbox}

The conclusion states the three assertions of Theorem~7.2 that are
uniform in the experiment: the score identity
\(P_\theta \ell_\theta = 0\), finiteness of the Fisher information,
and the in-probability quadratic expansion.  It carries no separate
openness or interior hypothesis: its parameter space is the typed
Euclidean space \verb|EuclideanSpace ℝ (Fin k)|, so local
perturbations \(\theta_0 + h_n/\sqrt n\) are well typed by
construction, with the global normalization carried by
\verb|IsPDFOf M μ|.  The remaining local content is the reusable
\verb|DifferentiableQuadraticMean M μ θ₀ ℓ|, whose
residual-integrability field the signature inherits
(Section~\ref{sec:concept-layer}).

The asymptotic normality \(\Delta_{n,\theta} \rightsquigarrow N(0,
I_\theta)\) lies outside this entry point: it does not depend on
quadratic-mean differentiability but is the classical multivariate
central limit theorem for the i.i.d.\ score vectors
\(\ell_\theta(X_i)\).  We therefore present only the three
differentiability-driven conclusions in \texttt{LAN\_expansion},
keeping the normality clause in the quoted statement above solely to
reproduce \cite[Theorem~7.2]{vanderVaart1998} in full.

\subsection{Asymptotic Representation Theorem}
\label{sec:art}

The Asymptotic Representation Theorem matches every weak limit of
statistics $T_n$ satisfying
$T_n \overset{h}{\rightsquigarrow} L_{\theta,h}$ under
$P^n_{\theta + h/\sqrt n}$ with a randomized statistic in the
Gaussian shift experiment.

\begin{theorem}[{\cite[Theorem~7.10]{vanderVaart1998}}]
Assume that the model is differentiable in quadratic mean at
$\theta$ with nonsingular Fisher information matrix $I_\theta$.
Let $T_n$ be statistics in the local experiments
$(P^n_{\theta+h/\sqrt n} : h \in \mathbb R^k)$ such that
$T_n$ converges in distribution under every $h$.  Then there
exists a randomized statistic $T$ in the normal experiment
$(N(h,I_\theta^{-1}) : h \in \mathbb R^k)$ such that $T_n$
converges in distribution to $T$ under every $h$.
\end{theorem}

Lean represents this randomized-statistic conclusion in the equivalent
Markov-kernel form: a randomized statistic based on a Gaussian observation
is equivalently a probability kernel from the Gaussian observation space to
the statistic space; this is the standard randomization formulation in
Le~Cam's comparison of experiments \cite[Chapter~2]{lecamYang2000}.

The Lean entry point is \texttt{LAN\_representation}; the signature is:
\begin{leanbox}
\small
\begin{verbatim}
theorem LAN_representation
    (M : ParametricFamily 𝓧 (Θ k)) (μ : Measure 𝓧) [SigmaFinite μ]
    (θ₀ : Θ k)
    (ℓ : 𝓧 → Θ k) (hℓ : Measurable ℓ)
    (hDQM : DifferentiableQuadraticMean M μ θ₀ ℓ)
    (J : Matrix (Fin k) (Fin k) ℝ) (hJ_pd : Matrix.PosDef J)
    (hJ : ∀ u v : Θ k, fisherInformation M μ θ₀ ℓ u v =
      ⟪u, (WithLp.equiv 2 _).symm (J.mulVec ((WithLp.equiv 2 _) v))⟫)
    (T : ∀ n, (Fin n → 𝓧) → 𝓨 d) (hT_meas : ∀ n, Measurable (T n))
    (L : Θ k → Measure (𝓨 d)) [∀ h, IsProbabilityMeasure (L h)]
    (hT_weak : ∀ h : Θ k,
      WeakConverges
        (fun n => (productMeasure M μ (θ₀ + (Real.sqrt n)⁻¹ • h) n).map (T n))
        (L h))
    [StandardBorelSpace (𝓨 d)] [Nonempty (𝓨 d)]
    [HasOuterApproxClosed (𝓨 d)] [BorelSpace (𝓨 d)]
    (hPDF : IsPDFOf M μ) :
    ∃ κ : Kernel (Θ k) (𝓨 d), IsMarkovKernel κ ∧
      ∀ h, L h = (multivariateGaussian h J⁻¹).bind κ
\end{verbatim}
\end{leanbox}

The Lean signature carries these assumptions as \texttt{lean-artifact}
witnesses rather than additional source-side restrictions.  The
measurability hypotheses \verb|hℓ| and \verb|hT_meas| expose the typed
data needed for score representatives and statistic pushforwards, while
\verb|hPDF| packages the density normalization and integrability facts
needed to form the product experiments.  The kernel conclusion also
requires descriptive-set-theoretic structure on the statistic space
(\verb|StandardBorelSpace|, \verb|Nonempty|, \verb|BorelSpace|, and
\verb|HasOuterApproxClosed|), automatic for $\mathbb R^d$ but explicit in
Lean.  Thus the extra rows in the signature record Lean's measure-kernel
encoding of Theorem~7.10, not stronger statistical assumptions.

\subsection{H\'ajek--Le~Cam Convolution Theorem}
\label{sec:convolution}

The Convolution Theorem decomposes the limit law of a regular
estimator as a Gaussian shift convolved with a noise factor, and
specializes to a covariance-matrix lower bound when the limit law
has finite second moment.

\begin{theorem}[{\cite[Theorem~8.8]{vanderVaart1998}}]
Assume that the experiment $(P_\theta : \theta \in \Theta)$ is
differentiable in quadratic mean at the point $\theta$ with
non-singular Fisher information matrix $I_\theta$.  Let $\psi$ be
differentiable at $\theta$.  Let $T_n$ be a regular estimator
sequence in the experiments $(P^n_\theta : \theta \in \Theta)$ with
limit distribution $L_\theta$.  Then there exists a probability
measure $M_\theta$ such that
\[
  L_\theta = N(0,\, \dot\psi_\theta I_\theta^{-1}
              \dot\psi_\theta^\top) \ast M_\theta .
\]
In particular, if $L_\theta$ has covariance matrix $\Sigma_\theta$,
then the matrix
$
  \Sigma_\theta - \dot\psi_\theta I_\theta^{-1}\dot\psi_\theta^\top
$
is non-negative-definite.
\end{theorem}

The Lean entry point chains LAN (Section~\ref{sec:lan}) and the
Asymptotic Representation Theorem (Section~\ref{sec:art}) through
an equivariance bridge.  The decomposition theorem itself has no
moment hypothesis:
\begin{leanbox}
\small
\begin{verbatim}
theorem hajek_le_cam_convolution_theorem
    (M : ParametricFamily 𝓧 (Θ k)) (μ : Measure 𝓧) [SigmaFinite μ]
    (θ₀ : Θ k)
    (ℓ : 𝓧 → Θ k) (hℓ : Measurable ℓ)
    (hDQM : DifferentiableQuadraticMean M μ θ₀ ℓ)
    (J : Matrix (Fin k) (Fin k) ℝ) (hJ_pd : Matrix.PosDef J)
    (hJ_fisher : ∀ u v : Θ k, fisherInformation M μ θ₀ ℓ u v =
      ⟪u, (WithLp.equiv 2 _).symm (J.mulVec ((WithLp.equiv 2 _) v))⟫)
    (ψ : Θ k → 𝓨 d) (ψDot : Θ k →L[ℝ] 𝓨 d)
    (hψ_diff : HasFDerivAt ψ ψDot θ₀)
    (ψDotMat : Matrix (Fin d) (Fin k) ℝ)
    (h_ψDot_mat : ∀ h : Θ k,
      ψDot h = (WithLp.equiv 2 _).symm (ψDotMat.mulVec ((WithLp.equiv 2 _) h)))
    (T : ∀ n, (Fin n → 𝓧) → 𝓨 d) (hT_meas : ∀ n, Measurable (T n))
    (hReg : RegularEstimatorSequence M μ θ₀ ψ T)
    (hPDF : IsPDFOf M μ) :
    ∃ M_θ : Measure (𝓨 d), IsProbabilityMeasure M_θ ∧
      hReg.limitDist =
        (ProbabilityTheory.multivariateGaussian (0 : 𝓨 d)
          (ψDotMat * J⁻¹ * ψDotMat.transpose)) ∗ M_θ
\end{verbatim}
\end{leanbox}
The covariance lower bound is packaged as a separate Lean entry
point, so the finite-second-moment witness is paid only by callers
asking for the covariance statement:
\begin{leanbox}
\small
\begin{verbatim}
theorem cov_psd_of_regular_estimator
    (M : ParametricFamily 𝓧 (Θ k)) (μ : Measure 𝓧) [SigmaFinite μ]
    (θ₀ : Θ k) (ℓ : 𝓧 → Θ k) (hℓ : Measurable ℓ)
    (hDQM : DifferentiableQuadraticMean M μ θ₀ ℓ)
    (J : Matrix (Fin k) (Fin k) ℝ) (hJ_pd : Matrix.PosDef J)
    (hJ_fisher : ∀ u v : Θ k, fisherInformation M μ θ₀ ℓ u v =
      ⟪u, (WithLp.equiv 2 _).symm (J.mulVec ((WithLp.equiv 2 _) v))⟫)
    (ψ : Θ k → 𝓨 d) (ψDot : Θ k →L[ℝ] 𝓨 d)
    (hψ_diff : HasFDerivAt ψ ψDot θ₀)
    (ψDotMat : Matrix (Fin d) (Fin k) ℝ)
    (h_ψDot_mat : ∀ h : Θ k,
      ψDot h = (WithLp.equiv 2 _).symm (ψDotMat.mulVec ((WithLp.equiv 2 _) h)))
    (T : ∀ n, (Fin n → 𝓧) → 𝓨 d) (hT_meas : ∀ n, Measurable (T n))
    (hReg : RegularEstimatorSequence M μ θ₀ ψ T)
    (hL_memLp : MemLp (fun y : 𝓨 d => y) 2 hReg.limitDist)
    (Sigmaθ : Matrix (Fin d) (Fin d) ℝ)
    (hSigmaθ_isCov : ∀ u v : 𝓨 d,
      ∫ y, ⟪u, y⟫ * ⟪v, y⟫ ∂hReg.limitDist
        - (∫ y, ⟪u, y⟫ ∂hReg.limitDist) * (∫ y, ⟪v, y⟫ ∂hReg.limitDist)
        = ⟪u, (WithLp.equiv 2 _).symm (Sigmaθ.mulVec ((WithLp.equiv 2 _) v))⟫)
    (hPDF : IsPDFOf M μ) :
    (Sigmaθ - ψDotMat * J⁻¹ * ψDotMat.transpose).PosSemidef
\end{verbatim}
\end{leanbox}

The Lean signature of the corollary carries the \texttt{implied} hypothesis: second-moment condition for the covariance positive semidefiniteness (PSD) conclusion.
\label{sec:hidden-moment}
The phrase ``if $L_\theta$ has covariance matrix $\Sigma_\theta$''
semantically presupposes $L_\theta$ to have finite second
moment: without it, $\Sigma_\theta$ is undefined and the
statement is vacuous.  The source text does not list the second
moment as a separate hypothesis, but the implication is
linguistic, not mathematical: ``has covariance matrix'' is
exactly ``has finite second moment plus the entries are the
$L^2$ inner products''.  Our Lean version scopes the
\verb|MemLp| witness as an internal antecedent to the covariance-PSD conclusion,
classified \texttt{implied} because the witnessing sentence is in
the source verbatim (see Section~\ref{sec:hidden-assumptions}).

\subsection{Local Asymptotic Minimax bound}
\label{sec:lam}

The local asymptotic minimax (LAM) bound gives a worst-case-over-finite-perturbations lower
bound on estimator risk; it is one of the most assumption-heavy cornerstones
on the parametric side and the one whose formalization surfaces the
sharpest theorem-vs-proof scope distinction.

\begin{theorem}[{\cite[Theorem~8.11]{vanderVaart1998}}]
Let the experiment $(P_\theta : \theta \in \Theta)$ be
differentiable in quadratic mean at $\theta$ with non-singular
Fisher information $I_\theta$.  Let $\psi$ be differentiable
at~$\theta$.  Let $T_n$ be any estimator sequence in the
experiments $(P^n_\theta : \theta \in \mathbb R^k)$.  Then for
any bowl-shaped loss function $\ell$,
\[
  \sup_{\substack{I \subset \mathbb R^k\\ \#I < \infty}}
   \liminf_{n \to \infty}\;
   \sup_{h \in I}\;
   \int \ell\!\left(\sqrt n (T_n - \psi(\theta + h/\sqrt n))\right)
   \,dP^n_{\theta + h/\sqrt n}
   \;\ge\;
   \int \ell\,dN(0,\, \dot\psi_\theta I_\theta^{-1}
                       \dot\psi_\theta^\top).
\]
\end{theorem}

Our Lean entry point is \texttt{local\_asymptotic\_minimax\_bound}.
Its proof reduces the theorem to
\texttt{localAsymptoticRisk\_ge\_target}, which combines with the Gaussian-shift minimax bound for
bowl-shaped losses.  The signature is:
\begin{leanbox}
\small
\begin{verbatim}
theorem local_asymptotic_minimax_bound
    (M : ParametricFamily 𝓧 (Θ k)) (μ : Measure 𝓧) [SigmaFinite μ]
    (θ₀ : Θ k)
    (ℓ : 𝓧 → Θ k) (hℓ : Measurable ℓ)
    (hDQM : DifferentiableQuadraticMean M μ θ₀ ℓ)
    (J : Matrix (Fin k) (Fin k) ℝ) (hJ : J.PosDef)
    (hJ_fisher : ∀ u v : Θ k, fisherInformation M μ θ₀ ℓ u v
      = ⟪u, (WithLp.equiv 2 _).symm (J.mulVec ((WithLp.equiv 2 _) v))⟫)
    (ψ : Θ k → 𝓨 d) (ψDot : Θ k →L[ℝ] 𝓨 d)
    (hψ_diff : HasFDerivAt ψ ψDot θ₀)
    (ψDotMat : Matrix (Fin d) (Fin k) ℝ)
    (h_ψDot_mat : ∀ h : Θ k,
      ψDot h = (WithLp.equiv 2 _).symm (ψDotMat.mulVec ((WithLp.equiv 2 _) h)))
    (T : ∀ n, (Fin n → 𝓧) → 𝓨 d) (hT_meas : ∀ n, Measurable (T n))
    (L : 𝓨 d → ℝ≥0∞)
    (hL_bowl : BowlShaped L) (hL_lsc : LowerSemicontinuous L)
    (hTight : MeasureTheory.IsTightMeasureSet
        (Set.range (fun n : ℕ =>
          (AsymptoticRepresentation.productMeasure M μ θ₀ n).map
            (fun ω => (Real.sqrt n) • (T n ω - ψ θ₀)))))
    (hPDF : IsPDFOf M μ) :
    localAsymptoticRisk M μ θ₀ T ψ L
      ≥ ∫⁻ y, L y ∂(multivariateGaussian (0 : 𝓨 d)
                    (ψDotMat * J⁻¹ * ψDotMat.transpose))
\end{verbatim}
\end{leanbox}

The proof in \cite[Theorem~8.11]{vanderVaart1998} restricts to
lower semicontinuous bowl-shaped $\ell$ and uniformly tight $T_n$.
These restrictions are not listed in the theorem header but appear
verbatim in the proof opener (``we only give the proof under the
further assumptions that the sequence $\sqrt n(T_n - \psi(\theta))$
is uniformly tight under $\theta$ and that $\ell$ is (lower)
semicontinuous'').  Our Lean signature promotes them to the typed
hypotheses \texttt{hL\_lsc} and \texttt{hTight}, so
\texttt{local\_asymptotic\_minimax\_bound} proves exactly the
restricted theorem the published proof justifies.  The row-level
audit classifies both as \texttt{high-confidence} with source
location in the proof body rather than the header.

\subsection{Semi-Parametric Convolution Theorem}
\label{sec:semi-parametric-cornerstone}

At the semi-parametric tier the convolution pattern reappears
under the lift from a Euclidean parameter
$\theta \in \mathbb R^k$ to an infinite-dimensional parameter
indexed by a tangent space
$\dot{\mathcal P}_P \subset L^2_0(P)$.

\begin{theorem}[{\cite[Theorem~25.20]{vanderVaart1998}}]
Let the function $\psi : \mathcal P \to \mathbb R^k$ be
differentiable at $P$ relative to the tangent cone
$\dot{\mathcal P}_P$ with efficient influence function
$\tilde\psi_P$.  Then the asymptotic covariance matrix of every
regular sequence of estimators is bounded below by
$P \tilde\psi_P \tilde\psi_P^\top$.  Furthermore, if
$\dot{\mathcal P}_P$ is a convex cone, then every limit
distribution $L$ of a regular sequence of estimators can be written
$L = N\!\bigl(0,\, P \tilde\psi_P \tilde\psi_P^\top\bigr) \ast M$ for
some probability distribution $M$.
\end{theorem}

We present the formalized theorem in its combined vector form below
($k \geq 1$); we write part~(a) for the covariance lower bound and
part~(b) for the convolution decomposition used by the Lean proof.
The displayed combined form
packages parts~(a) and~(b), so it carries the finite-second-moment
witness needed for part~(a); the part-specific entry point for
part~(b) has the same common inputs but omits that moment witness.

\begin{leanbox}
\small
\begin{verbatim}
theorem semiparametric_convolution_theorem_vec
    {ψ : Measure Ω → EuclideanSpace ℝ (Fin k)}
    (hψ : PathwiseDifferentiableAt_vec P (tangentSpace T_set) ψ)
    {IF_eff : Fin k → ↥(L2ZeroMean P)}
    (hEIF : IsEfficientInfluenceFunction_vec hψ.derivative IF_eff)
    (T_n : ∀ n, (Fin n → Ω) → EuclideanSpace ℝ (Fin k))
    (hT_meas : ∀ n, Measurable (T_n n))
    (L : Measure (EuclideanSpace ℝ (Fin k))) [IsProbabilityMeasure L]
    (hReg : IsRegularEstimator_vec P T_set ψ hψ hEIF T_n L)
    (hL_memLp : MemLp (id : EuclideanSpace ℝ (Fin k) → _) 2 L) :
    let G : Matrix (Fin k) (Fin k) ℝ := Matrix.gram ℝ IF_eff
    let Sigma : Matrix (Fin k) (Fin k) ℝ := fun i j =>
      ∫ y, (y.ofLp i - ∫ z, z.ofLp i ∂L)
          * (y.ofLp j - ∫ z, z.ofLp j ∂L) ∂L
    (Sigma - G).PosSemidef ∧
    (∃ M : Measure (EuclideanSpace ℝ (Fin k)), IsProbabilityMeasure M ∧
      L = (ProbabilityTheory.multivariateGaussian
            (0 : EuclideanSpace ℝ (Fin k)) G) ∗ M)
\end{verbatim}
\end{leanbox}

The tangent-space encoding separates two roles.  The pathwise
derivative and efficient influence functions live over
\texttt{tangentSpace T\_set}, the $L^2$-closed span of
the user-supplied tangent set; this is the Hilbert-space object used
for projections, Gram matrices, and efficient-influence-function
approximation.  The regular-estimator hypothesis, however, quantifies
score directions only over the algebraic span
\verb|Submodule.span ℝ T_set.carrier|, matching
\cite[p.~366]{vanderVaart1998} (``for every $g \in \mathrm{lin}\, g_p$'').
The released theorem formalizes the linear-space case of
Theorem~25.20 in \cite{vanderVaart1998}, which the source proof establishes first (``Assume
first that the tangent set is a linear space''), by applying the
finite-dimensional argument on algebraic-span subspaces and passing to
the closed-span limit through the efficient-influence-function
approximation.  In the displayed vector headline, the proof works over
the closed linear span generated by \verb|T_set.carrier|; this space is
automatically a convex cone, so no separate cone hypothesis appears in
that vector signature.
The proof reuses the LAN expansion (Section~\ref{sec:lan}) and
the Asymptotic Representation Theorem (Section~\ref{sec:art}), and
applies the H\'ajek--Le~Cam Convolution Theorem (Section~\ref{sec:convolution}) to the
sigmoid-family submodels of \cite[Example~25.16]{vanderVaart1998}.
This route replaces an earlier closure through the same theorem's
Fr\'echet wrapper, which forced a uniform-Hadamard remainder
condition; the architectural choices behind the rewrite (including
the three drift hypotheses it eliminated) are the subject of the
case study in Section~\ref{sec:eval-semipar-convolution-drift}.

\subsection{Selected Surfaced Hypotheses by Tier}
\label{sec:hidden-assumptions}

The cornerstone formalizations above surface hypotheses that a
reader of the source prose might not expect to track explicitly.
We classify the cornerstone-by-cornerstone instances using the
scheme of Section~\ref{sec:auditor}.  The list below is selective:
it collects the recurring or load-bearing rows used in the main-text
discussion; a reference-document fragment appears in
Appendix~\ref{app:book-reference-example}.
The bullets name the Lean object, point to the relevant source
evidence when applicable, and state why the assigned tier applies.

\noindent \textbf{Hypotheses labeled with }\texttt{high-confidence}.
\begin{itemize}\itemsep1pt
  \item \verb|hL_lsc| (lower semicontinuity of the loss, LAM
    bound): \cite[Chapter~8]{vanderVaart1998}, proof
    opener of Theorem~8.11 verbatim.
  \item \verb|hTight| (uniform tightness of the recentered
    estimator, LAM bound): same proof opener
    (Section~\ref{sec:lam}).
  \item \verb|tangentSpace| (closed linear span, semi-parametric
    Convolution Theorem): takes the closure of the user-supplied span,
    matching
    \cite[Chapter~25.3]{vanderVaart1998} verbatim; the efficient
    influence function is quantified over this closed span, and the
    convolution-decomposition statement uses it through the EIF Gram
    matrix.
\end{itemize}

\noindent \textbf{Hypotheses labeled with }\texttt{lean-artifact}.
\begin{itemize}\itemsep1pt
  \item Parametric \verb|DifferentiableQuadraticMean| residual
    \verb|mem| field: Bochner-integral representation choice
    for the parametric LAN cornerstone's real-integral form
    (Lean's $\int f\,d\mu$ returns $0$ on non-integrable $f$,
    whereas the source expression $\int f^2\,d\mu < \infty$
    presupposes integrability).
  \item Measurability witnesses such as \verb|hℓ| and
    \verb|hT_meas|: Lean requires them when forming integrals, product
    laws, and pushforwards from the score representatives and estimator
    statistics of the source setup (Section~\ref{sec:art}).
  \item Density-family adapter \verb|hPDF : IsPDFOf M μ|: it
    writes $p_\theta$ as a density of $P_\theta$ with respect to
    a dominating measure \cite{vanderVaart1998}; the Lean API packages the corresponding
    normalization, nonnegativity, and integrability facts explicitly.
\end{itemize}

\noindent \textbf{Hypotheses labeled with }\texttt{implied}.
\begin{itemize}\itemsep1pt
  \item \verb|hL_memLp| for covariance-matrix conclusions.  In
    the H\'ajek--Le~Cam Convolution Theorem, the witnessing
    sentence is \cite[p.~115]{vanderVaart1998} ``if $L_\theta$
    has covariance matrix''; in the semi-parametric Convolution
    Theorem, the statement uses the phrase ``asymptotic covariance
    matrix''.  Both phrases semantically presuppose a finite second
    moment, so the Lean moment witness is scoped only to the
    covariance-PSD/covariance-Gram conclusions
    (Sections~\ref{sec:hidden-moment} and
    \ref{sec:semi-parametric-cornerstone}).
\end{itemize}

\noindent \textbf{No hypotheses labeled with }\texttt{potential-drift}.
On the cornerstone chains this column is empty: every
hypothesis on a cornerstone main signature either matches a
reference row directly or has been reverted.  The operational
referent is the LAM-bound drift episode of
Section~\ref{sec:eval-lam-drift}: its clearest row was
\verb|hL_coercive|, a coercivity strengthening of the source
bowl-shaped-loss assumption, but the same audit also reverted
\verb|hL_continuous|, \verb|h_L_orig_avgRisk_bd|, and
\verb|hψ_meas|.  Thus the empty \texttt{potential-drift} column
records the post-revert theorem state;
Section~\ref{sec:eval-lam-drift} describes how the drift was introduced
and removed.

\section{Conclusion}
\label{sec:conclusion}

We have presented a systematic Lean~4 formalization of asymptotic
statistical estimation theory, covering core parametric and
semi-parametric limit theorems through reusable probability
infrastructure, statistical concept definitions, and theorem-level
entry points.  The accompanying multi-agent scaffold addresses the
thin-Mathlib-coverage setting of this domain, while the
hypothesis-disciplined audit makes source-faithfulness a runtime
obligation rather than a consequence of compilation alone.  Together,
the formal library and the audit methodology show how agent-assisted
formalization can scale while keeping theorem assumptions tied to the
mathematical source.

Future work has two immediate directions.  On the mathematical side,
the library should be extended toward broader asymptotic statistics,
including additional limit experiments, efficiency results, and
nonparametric settings.  On the methodological side, the audit should
be tested across other assumption-heavy domains and agent platforms,
with more automation for extracting candidate theorem boundaries and
tracking source evidence across refactors.

\bibliographystyle{plain}
\bibliography{references}

\appendix

\section{Evaluation}
\label{sec:evaluation}

In this appendix, together with the two case studies presented in
Appendix~\ref{app:case-studies}, we answer three questions about the
methodological contributions of Section~\ref{sec:contributions}: (i) is the closed
artifact axiom-clean at the scale claimed
(Section~\ref{sec:eval-scale}); (ii) do the row-level signature audit
and the definition audit catch failure modes that \verb|lake build|,
\verb|#print axioms|, and subjective mathematical review let pass
(Section~\ref{sec:eval-audit}); and (iii) do the two case studies
exhibit difficulties specific to faithful statistical
formalization that stress-test the audit discipline introduced in
Section~\ref{sec:hypothesis-discipline}
(Appendix~\ref{app:case-studies}).  The LAM bound isolates the
failure mode at the signature layer; the semi-parametric Convolution
Theorem extends it to the definition layer.  Operational metrics
(commit attribution, retrieval usage, wave terminal-state breakdown,
cost concentration, and the full falsification log) are treated as
project-run artifacts rather than load-bearing evidence for the
released theorem statements.

\subsection{Platform Requirements}
\label{app:platform}

The framework does not require a custom agent-platform integration,
but it does assume a small set of harness primitives:
\begin{enumerate}\itemsep1pt
  \item Agent dispatch into an isolated execution context.  In
    our case a git worktree on the same host, with the Lake package
    directory symlinked and the build cache copied for fast rebuilds.
  \item Durable task cards and self-identified terminal-state
    returns for each dispatched agent (the recorded wave outcomes of
    Table~\ref{tab:terminal-states}).
  \item Tool-mediated build verification through the platform's
    shell primitive (\texttt{lake build} invoked from inside the
    worktree by both Executors and the gating Reviewer).
\end{enumerate}
These primitives are common in current coding-agent environments,
but this paper evaluates only one implementation: every experiment
used Claude Code (Opus~4.7) for both the Manager session and all
agent dispatches.  We have not validated the framework end-to-end on
a platform other than Claude Code.

\subsection{Scale and Axiom Certificates}
\label{sec:eval-scale}

The released library comprises about 80{,}700 lines of Lean~4 code
across 151 files: roughly 240 definitional declarations and about a
thousand lemmas and theorems (counted by a line-prefix scan, so
indicative rather than exact), so the definitional surface is compact
relative to the proof effort built on it.  Every declaration in the
released library is axiom-clean, with no \verb|sorryAx|;
Appendix~\ref{app:axioms} records the per-cornerstone certificates and
the exact baseline axiom set.

The library can support further formalization in the same domain: a
substantial part of it is general-purpose probability, measure-theory,
and analysis material that fills gaps in current Mathlib and is a
candidate for upstreaming, and this reusable infrastructure is
distributed across the library rather than confined to any one
directory.  The cornerstone entry-point theorems sit above it as
axiom-clean, named entry points.

\subsection{Hypothesis Classification and Audit Outcomes}
\label{sec:eval-audit}

Across the five released cornerstones, the audit layer turns
source-faithfulness into a row-level property of theorem and
definition boundaries.  Each cornerstone's reference document
classifies every signature hypothesis and instance binder into the
four tiers of Section~\ref{sec:hypothesis-discipline}, and the
load-bearing outcome is uniform across the five released cornerstones:
\textbf{no cornerstone chain carries a \texttt{potential-drift} row}.
The only \texttt{implied} rows are the finite-second-moment
prerequisites for covariance or variance statements.  The mechanical audit script reports
\textsc{Mechanical checks: PASS} on all five cornerstones, and each row
is tied to a source quotation, an implied mathematical prerequisite,
or a concrete Lean encoding correspondence.

These rows make the extra structure in the Lean statements
accountable.  A reviewer can inspect, argument by argument, whether
a formal boundary is source-stated, source-implied, required by the
encoding, or unsupported.  Appendix~\ref{app:case-studies} gives the
two case studies behind this claim: first at the main-signature
boundary of the LAM theorem, and then at the definition boundary of
the semi-parametric Convolution Theorem.

\section{Case Studies and Supporting Evidence}
\label{app:case-studies}

\subsection{Case Study: The LAM-Bound Drift Episode}
\label{sec:eval-lam-drift}

The LAM bound shows the control loop at the theorem-signature
boundary.  An initially plausible route produced a closed but
over-strengthened statement; the audit rows exposed the unsupported
assumptions, and the Manager sent the theorem back to planning rather
than accepting the stronger boundary.  The released LAM theorem is
closed with no extra unsupported assumptions.

\noindent \textbf{The setting.}
Early in LAM development the proof tried to shortcut the finite-prior
LAN bridge by moving directly to Gaussian-$\tau$-prior average-risk
objects at the sub-lemma level.  That shortcut mixed the finite-prior
comparison step with the later Gaussian-shift minimax calculation, so
the intermediate hypotheses no longer had a clean anchor in the
reference text.  The skeleton was build-clean (\verb|lake build|
succeeded), and subsequent dispatch batches chained off it.

\noindent \textbf{Drift accumulation.}
During the body phase, proof blockers led later route revisions to
propose four unsupported assumptions on candidate LAM signatures:
\verb|hL_coercive|, \verb|hL_continuous|,
\verb|h_L_orig_avgRisk_bd|, and \verb|hψ_meas|.  Each was
introduced to discharge a blocker raised by the mis-specified
collapse, strengthening the hypothesis set rather than questioning the
strategy.  The result was an axiom-clean, \verb|sorry|-free chain
with a strictly over-strengthened hypothesis boundary, and hence a
weaker theorem than the standard LAM statement: the constant-zero loss
function is bowl-shaped but not coercive, refuting
\verb|hL_coercive| as a hypothesis a faithful formalization could
carry.  The other rows failed the audit by
strengthening the theorem boundary or imposing a condition outside the
source theorem's scope.

The measurability row is a useful inversion of the usual source-text
problem.  Human prose often leaves measurability obligations implicit,
but here the agent-side pressure went in the opposite direction:
blocked routes promoted a global measurability assumption on the
functional, \verb|hψ_meas|, to the theorem boundary even though the LAM
theorem does not require it there.  The audit therefore treated
\verb|hψ_meas| as a route artifact, not a released assumption.

\noindent \textbf{Route correction and released state.}
The row-level reference made the strengthened boundary visible:
\verb|hL_coercive| strictly strengthens the ``bowl-shaped''
hypothesis of \cite[Theorem~8.11]{vanderVaart1998},
\verb|hL_continuous| strictly strengthens the proof opener's
lower-semicontinuity condition, \verb|hψ_meas| has no source anchor,
and \verb|h_L_orig_avgRisk_bd| restricts estimators beyond the
source's regularity.  All four were therefore recorded as
\texttt{potential-drift}.  Under the workflow of
Section~\ref{sec:orchestration}, the Manager treated these rows as route
errors rather than accepted hypotheses, sending the theorem back to
the Planner.  The replacement proof follows the source argument more
closely.  In the Lean development, the route is exposed by
\texttt{localAsymptoticRisk\_\allowbreak ge\_\allowbreak target}: it first passes to a
subsequential Gaussian-shift limit for the recentered
estimator--score pair, then transfers lower bounds from rational local
parameters to all local parameters by lower semicontinuity, and finally
applies the Gaussian-shift bowl-shaped-loss lower bound.

The released LAM statement is
therefore not the strengthened intermediate theorem: its audited
boundary has no \texttt{potential-drift} row, and every remaining
formal argument is source-stated, source-implied, or justified by a
concrete Lean encoding correspondence.

The episode is the operational referent of the theorem-boundary claim:
the workflow converted four unsupported assumptions into a route
change and finished the theorem without carrying them in the released
signature.

\subsection{Case Study: The Semi-Parametric Convolution Theorem}
\label{sec:eval-semipar-convolution-drift}

The semi-parametric Convolution Theorem
(Section~\ref{sec:semi-parametric-cornerstone}) shows the same
control loop one layer deeper, at the definition boundary.  In one
development cycle, a second-moment prerequisite is accepted as
\texttt{implied}, three signature-side drifts are rejected as
\texttt{potential-drift}, an attempted relocation into authored
structure fields is made visible by the definition audit, and the
final proof is recovered by a strategic rewrite rather than by adding
hypotheses.  The released theorem and audited definitions have no
extra unsupported assumptions: every non-verbatim input is either
classified as \texttt{implied} or justified as a Lean representation
choice.

\noindent \textbf{The implied second-moment assumption.}
The source theorem states part~(a) as ``the asymptotic
covariance matrix of every regular sequence of estimators
is bounded below by $P \tilde\psi_P \tilde\psi_P^\top$''; the
phrase ``covariance matrix'' presupposes the matrix exists as
$\mathbb E[(X - \mathbb E X)(X - \mathbb E X)^\top]$, requiring
$\mathbb E[X_i^2] < \infty$, i.e.\ a finite second moment of $L$
that the source theorem does not state separately.

The first attempt to close part~(a) blocked at the covariance lower
bound.  In the released Lean route, the finite-second-moment fact is
not derived from the other regularity hypotheses; it is supplied
explicitly because the source phrase ``covariance matrix'' already
presupposes such a moment condition.  Without this \texttt{MemLp 2}
witness, the encoded covariance inequality does not express the
intended lower bound through Mathlib's covariance and moment APIs.  Re-reading
against the row-level reference identified ``covariance matrix'' as the
source; the finite-second-moment hypothesis was classified
\texttt{implied} (the source phrase ``covariance matrix'' is the
witnessing sentence).  A subsequent split of the Lean entry points
(Section~\ref{sec:semi-parametric-cornerstone}) confines it to
part~(a); part~(b)'s decomposition is well-defined for arbitrary $L$.

\noindent \textbf{Drift accumulation, made visible by the row-level reference.}
Independently of this implied-hypothesis surfacing, the original
closure of the cornerstone routed through the parametric
Convolution Theorem's Fr\'echet wrapper
(Section~\ref{sec:convolution}), which forced uniform Hadamard.  As
the proof route developed, three unsupported assumptions appeared as
route pressure: a
bounded-density regularity condition on the tangent set
(\verb|hT_dense|); a uniform-Hadamard remainder bound strictly
stronger than the source's pointwise pathwise-differentiability
(\verb|hψ_Hadamard_remainder|); and an externally supplied path
family with its score-equation hypothesis
(\verb|γ + hγ_score|), where the source proof constructs the family
inside the proof body.

The bounded-density and uniform-Hadamard assumptions reached the
main theorem signature, while the path-family hypothesis appeared only
on an intermediate candidate route.  None matched a source row, so all
three were classified \texttt{potential-drift}.  As in the LAM
case, that classification did not authorize a stronger theorem; it
returned the route to planning.

\noindent \textbf{Failed relocation into definitions.}
A later route tried to move the same conditions one layer down, into
authored structure fields.  It added a \verb|hadamard_remainder| field to
\leanPathwiseDifferentiableAt{} and a \verb|bounded_dense| field
to \verb|TangentSpec|, dropping the two corresponding hypotheses
from the main signature; the \texttt{potential-drift} column on the
theorem momentarily emptied, the build stayed green, and
\verb|#print axioms| stayed at the project baseline.

The definition audit made the relocation visible.  The added fields
had no source anchor: the source tangent-set definition
(Chapter~25.3) carries no bounded-density regularity condition, and
the pathwise-differentiability definition
(\cite[p.~363]{vanderVaart1998}) is pointwise, not
uniform-Hadamard.  Both fields were reverted, and the two hypotheses
returned to the \texttt{potential-drift} column.  A parallel
definition-side episode on the \verb|QMDPath| structure
(\verb|qmd_residual_memLp| added as a separate field, then removed by
restating the quadratic-mean limit in
$\overline{\mathbb R}_{\ge 0}$ form) is documented in
Appendix~\ref{app:phase-sigmoid-drift-sequence}.  Across both
episodes, the system forced drift to surface either as a theorem-side
\texttt{potential-drift} row or as a definition-side
\texttt{potential-drift} row, leaving no path through silent
definitional strengthening.

\noindent \textbf{Strategic rewrite and released state.}
Once the audit had rejected both repairs---adding the assumptions to
the theorem signature and relocating them into authored structure
fields---the original proof route had no faithful local patch.  The
remaining option was an architectural rewrite.  Phase Sigmoid
re-architected the proof chain around \cite[Example~25.16]{vanderVaart1998}
(construction in Section~\ref{sec:semi-parametric-cornerstone}),
eliminating all three drift pressures at the source.  No new
structure field, no moment strengthening on the basis scores; the
post-Phase-Sigmoid \texttt{potential-drift} and definition-side \texttt{potential-drift} columns are both empty.
The definition-level encoding of \verb|IsRegularEstimator| and the
algebraic-span scope decision are addressed in
Appendix~\ref{app:cornerstone-defs}.

\subsection{Semi-Parametric Reference Rows and Drift Sequence}
\label{app:book-reference-example}
\label{app:phase-sigmoid-drift-sequence}

Table~\ref{tab:book-reference-semipar-convolution-abridged} reproduces
representative rows from the book-reference document shipped with the
semi-parametric Convolution Theorem artifact
(\cite[Theorem~25.20]{vanderVaart1998}), focusing on rows relevant to
the convolution-decomposition part (part~(b)).

\begin{table}[t]
  \centering
  \small
  \renewcommand{\arraystretch}{1.05}
  \setlength{\tabcolsep}{3pt}
  \begin{tabularx}{\linewidth}{@{}>{\raggedright\arraybackslash}p{0.17\linewidth}
      >{\raggedright\arraybackslash}p{0.14\linewidth} X@{}}
    \toprule
    \textbf{Row} & \textbf{Ref.} & \textbf{Verbatim opener / justification} \\
    \midrule
    \multicolumn{3}{@{}l}{\textbf{\texttt{high-confidence}}} \\
    \addlinespace[2pt]
    \verb|T_set|
      & Ch.~25.3 / p.\,366
      & \emph{``\,\ldots\ differentiable at $P$ \textbf{relative to the tangent cone $\dot{\mathcal P}_P$}\,\ldots''}  Tangent cone source: \cite[Chapter~25.3, p.\,362]{vanderVaart1998}. \\
    \addlinespace[2pt]
    \verb|hReg|
      & Ch.~25.3 / pp.\,365--366
      & \emph{``An estimator sequence $T_n$ is called \textbf{regular at $P$} for estimating $\psi(P)$ \ldots''}  Encoded over the algebraic span \verb|Submodule.span ℝ T_set.carrier| (vdV's ``$g \in \mathrm{lin}\, g_p$,'' p.~366). \\
    \midrule
    \multicolumn{3}{@{}l}{\textbf{\texttt{lean-artifact}}} \\
    \addlinespace[2pt]
    \verb|hT_meas|
      & Ch.~25.3 / p.\,365
      & Estimator measurability, stated verbatim in the source: \emph{``as usual, an estimator sequence $T_n$ is a measurable function $T_n(X_1, \ldots, X_n)$ of the observations''} \cite[Chapter~25.3, p.\,365]{vanderVaart1998}.  \verb|hT_meas| is the explicit Lean witness of this source-stated property, used to form pushforwards and integrals. \\
    \midrule
    \multicolumn{3}{@{}l}{\textbf{\texttt{potential-drift} (historical)}} \\
    \addlinespace[2pt]
    \verb|hT_dense|
      & none
      & \textbf{[Eliminated]} Bounded-density regularity on the tangent set, demanded by an early character-function step; removed by the Phase Sigmoid rewrite. \\
    \addlinespace[2pt]
    \texttt{Hadamard rem.}
      & none
      & \textbf{[Eliminated]} A uniform-Hadamard rate, strictly stronger than the source's pointwise pathwise-differentiability (\cite[p.~363]{vanderVaart1998}); removed by the same rewrite. \\
    \midrule
    \multicolumn{3}{@{}l}{\textbf{\texttt{high-confidence} definition-side row}} \\
    \addlinespace[2pt]
    \texttt{IsRegular}\newline\texttt{Estimator\_vec}
      & Ch.~25.3 / p.\,365
      & \emph{``An estimator sequence $T_n$ is called \textbf{regular at $P$} for estimating $\psi(P)$\,\ldots\ for every $g \in \dot{\mathcal P}_P$.''}  This is the source anchor for the chosen-submodel regularity pattern: one realizing submodel is selected for each score direction in the algebraic-span scope \verb|Submodule.span ℝ T_set.carrier| used by the released theorem (vdV's ``$g \in \mathrm{lin}\, g_p$''). \\
    \midrule
    \multicolumn{3}{@{}l}{\textbf{reverted definition-side drift (historical)}} \\
    \addlinespace[2pt]
    \texttt{Hadamard rem.}
      & none
      & \textbf{[Reverted]} Field added to the pathwise-differentiability structure to hide the same uniform-Hadamard requirement; the definition audit caught that \cite[p.~363]{vanderVaart1998} is pointwise, with no uniform rate. \\
    \addlinespace[2pt]
  \texttt{bounded\_dense}
      & none
      & \textbf{[Reverted]} Field added to the tangent-set definition; \cite[Chapter~25.3]{vanderVaart1998} defines the tangent set as a collection of score functions in $L^2(P)$ arising from QMD submodels, with no bounded-density condition, so the field was removed. \\
    \bottomrule
  \end{tabularx}
\caption{Abridged excerpt from the book-reference document
    accompanying the semi-parametric Convolution Theorem
    cornerstone.  The table is a selected excerpt rather than the
    complete audit record.  \texttt{potential-drift} is empty on the
    released artifact; all \texttt{potential-drift} rows shown here are
    historical, kept for parallel with the development case study
    (Section~\ref{sec:eval-semipar-convolution-drift}) and marked
    \textbf{[Eliminated]} or \textbf{[Reverted]} accordingly.}
  \label{tab:book-reference-semipar-convolution-abridged}
\end{table}

The historical rows in the table correspond to the following
development sequence, expanding the compressed summary in
Section~\ref{sec:eval-semipar-convolution-drift}: the wave-by-wave
drift accumulation, the attempted relocation into authored structure
fields, and the architectural rewrite that eliminated all three.

\begin{itemize}\itemsep1pt
  \item \textbf{First drift wave}: agent introduces
    \verb|hT_dense| (a bounded-density regularity condition on
    the tangent set) to the main signature, demanded by an
    early character-function deriver closure step.  Not in the
    tangent-set definition of
    \cite[Chapter~25.3]{vanderVaart1998}.
  \item \textbf{Second drift wave}: \verb|hψ_Hadamard_remainder|
    added: a uniform-Hadamard remainder strictly stronger than
    the source's pointwise pathwise-differentiability (\cite[p.~363]{vanderVaart1998}).
  \item \textbf{Third drift wave}: a
    user-supplied path family \verb|γ| together with its
    score-equation hypothesis \verb|hγ_score|, where the source proof
    constructs the family \textbf{inside} the proof body rather
    than receiving it as a hypothesis.  This third drift appeared only
    on an intermediate candidate route.
  \item After three waves, two unsupported hypotheses had reached the
    main theorem signature, and a third had appeared on that
    intermediate route; \verb|lake build| was green, \verb|#print axioms|
    stayed at the project baseline, and there was no \verb|sorry|.  The
    row-level reference audit flagged all three as non-matching any
    \texttt{high-confidence} row.
  \item \textbf{Attempted relocation into definitions}: rather
    than rewrite the proof, the same conditions are moved one
    layer down: a \verb|hadamard_remainder| field added to
    \leanPathwiseDifferentiableAt{}, and a \verb|bounded_dense|
    field added to \verb|TangentSpec|.  The main signature's
    \texttt{potential-drift} column momentarily empties; \verb|lake build| stays
    green, \verb|#print axioms| stays at the baseline.
  \item \textbf{Definition audit blocks the relocation}:
    the added fields have no source anchor: \cite[Chapter~25.3]{vanderVaart1998}
    defines the tangent set as a collection of score functions in
    $L^2(P)$ arising from QMD submodels, with no bounded-density
    condition, and \cite[p.~363]{vanderVaart1998} defines pathwise
    differentiability as a pointwise statement along each
    submodel, with no uniform Hadamard rate.  Both fields are
    reverted in a single commit; the two signature-side drift rows
    return to the \texttt{potential-drift} column.
  \item \textbf{Parallel definition-side episode}: a \verb|qmd_residual_memLp|
    field is caught on the \verb|QMDPath| structure (an
    $L^2$-membership claim on the DQM residual, absent from
    the source's Chapter~25.3 eq.~(25.13)).  This is the
    semi-parametric \verb|QMDPath| structure, distinct from the
    parametric \verb|DifferentiableQuadraticMean| residual \verb|mem|
    field discussed in Section~\ref{sec:hidden-assumptions}: the
    parametric cornerstone keeps vdV's real-integral equation~(7.1),
    where the extra field restores the Lebesgue-integrability
    content, whereas this semi-parametric cornerstone adopts an
    $\overline{\mathbb R}_{\ge 0}$ \verb|eLpNorm| form of
    equation~(25.13).  Accordingly, the added field is removed by restating the
    quadratic-mean limit in $\overline{\mathbb R}_{\ge 0}$ form
    and recovering the previous corollary statement as a
    downstream lemma rather than a structure field.
  \item \textbf{Architectural rewrite (Phase Sigmoid)}: the
    proof chain is re-architected around \cite[Example~25.16]{vanderVaart1998}'s
    sigmoid-family construction, eliminating all three drift
    pressures at the source.  No new structure field, no moment
    strengthening on the basis scores.  The audit script applied
    to the post-rewrite state confirms no signature-side and no
    definition-side drift.
\end{itemize}

\subsection{Joint Lesson}
\label{sec:eval-joint}

The LAM case study (Section~\ref{sec:eval-lam-drift}) establishes
the theorem-boundary layer: unsupported assumptions on a main
signature are routed back into planning until the released theorem
has none.  The semi-parametric Convolution Theorem extends this in
two directions.  First, when the same pressure moves into authored
structure fields, the definition audit makes the relocation visible
and returns it to the same source-anchor discipline.  Second, when
both signature and definition boundaries are clean but the proof
route still demands unsupported assumptions, the resolution is
architectural, not a local patch.  Phase Sigmoid is the
operational referent: it replaces the proof route rather than
adding another hypothesis or definition field.

\section{Verification Artifacts and Lean-Side Details}
\label{app:verification}

This appendix collects the Lean-side details that support the domain
claims of Section~\ref{sec:domain} --- axiom certificates, signature
pinning, and the load-bearing concept definitions that the cornerstone
signatures are stated in terms of --- to keep the main text
focused on the conceptual contributions.

\subsection{Axiom Certificates}
\label{app:axioms}

Each of the five cornerstones (LAN expansion, Asymptotic Representation, H\'ajek--Le~Cam Convolution, Local Asymptotic Minimax bound, and the semi-parametric Convolution Theorem (Section~\ref{sec:semi-parametric-cornerstone})), including the
covariance-PSD conclusion associated with H\'ajek--Le~Cam,
compiles under the same three baseline axioms.
Running \verb|#print axioms| on each top-level declaration produces

\begin{verbatim}
[propext, Classical.choice, Quot.sound]
\end{verbatim}

\noindent with no \verb|sorryAx|.  Thus every step on the main
chain of each theorem is closed by typed Lean terms rather than
deferred proof placeholders.  The released artifact contains no
remaining \verb|sorry|, on-chain or off-chain.

\subsection{Signature Pinning}
\label{app:signature-pinning}

The axiom certificate of Appendix~\ref{app:axioms} guarantees that every
cornerstone is closed by typed Lean terms; it says nothing about what those
terms claim. A build can stay green, sorry-free, and axiom-clean while a
hypothesis, an instance binder, the conclusion, or an authored definition body
silently changes meaning. The signature-pinning pass closes this gap on the
mechanical side. It records the fully elaborated signature of each cornerstone
and the body of each authored definition the cornerstones depend on
as a committed snapshot, and a checker
fails whenever a regenerated snapshot differs. The mechanical layer therefore
certifies two complementary facts: that each theorem is well typed (the axiom
certificate) and that no tracked signature element changes without surfacing in
a single snapshot comparison (signature pinning). Faithfulness to \cite{vanderVaart1998}, the judgment
that each surfaced element matches the source, remains the human-and-model row
audit over the book-reference document
(Section~\ref{sec:hypothesis-discipline}).

On the type side, membership is computed rather than curated. The tracked roots
are exactly the
cornerstone theorems this paper claims; printing each elaborated type pins, for
free, every instance binder, universe level, and quantifier the type mentions,
so no instance argument is selected by hand; the only human choice is which
cornerstones to pin, already fixed by the paper's scope. A definition the type names is
pinned there only by use; the body that gives it meaning is tracked separately.
The tracked bodies
are the authored book-concept definitions the cornerstone signatures are stated
in terms of (Appendix~\ref{app:cornerstone-defs}). A body is pinned because changing it
relocates a proof obligation into a definition (the definition-drift channel of
Section~\ref{sec:hidden-assumptions}) without touching any hypothesis. Mathlib
definitions, including \verb|multivariateGaussian|, are trusted base and are not
body-tracked; the pinned toolchain guards them.

The signature pass of the audit suite (Section~\ref{sec:audit}) reconciles a
signature's explicit hypotheses against their reference rows by name and count.
Signature pinning is what extends mechanical coverage to the instance arguments,
the conclusion, and the definition bodies: these enter through the elaborated
type and the printed body rather than through the name-and-count reconciliation.
The snapshot is taken with universe levels shown but proofs and notation left
folded. Full expansion of every instance and operator is faithful but yields an
unreadable artifact (on the order of $10^5$ lines), whereas the universe-level
form is under a thousand lines and remains a human-readable review surface;
proof-term arguments are elided by proof irrelevance, so an unrelated proof edit
does not perturb the snapshot. The single channel left uncovered is an internal
instance replacement that prints identically, which the row audit covers
semantically on the affected hypothesis.

Because the toolchain is pinned, the snapshot moves only when a tracked
signature changes or the toolchain is deliberately advanced, the latter
regenerated within the same change. A coupling check rejects any commit
that stages a new snapshot without staging a book-reference update alongside it,
so a signature change cannot land without surfacing to the faithfulness audit in
the same commit.

\subsection{Cornerstone-Load-Bearing Definitions}
\label{app:cornerstone-defs}

The theorem signatures in Section~\ref{sec:domain} are stated in
terms of authored \texttt{structure}s and \texttt{def}s whose fields
carry source content not visible on the signature alone.  The case
studies of
Section~\ref{sec:eval-semipar-convolution-drift} (the relocated \verb|hadamard_remainder|
and \verb|bounded_dense| fields, and the rejected
\verb|qmd_residual_memLp| field) hinge on these field layouts: the
definition audit asks whether each structure field has a source
anchor, and the verdict is over the \textbf{fields}, not the
structure's name.  The main text already expands the DQM, regular
estimator, and tangent-set definitions
(Section~\ref{sec:concept-layer}); we do not repeat those definitions
here.  The supplementary source excerpts below are a readability-oriented selection of
load-bearing declarations needed to read the cornerstone signatures
and the case-study discussion; they are not the complete tracked set of
Appendix~\ref{app:signature-pinning}.  The displayed declarations are source
excerpts under their file-level namespace and variable context.

\noindent \textbf{Vector-valued pathwise differentiability (\leanPathwiseDifferentiableAtVec).}

\noindent \cite[Chapter~25.3, pp.~362--363]{vanderVaart1998}.  A vector functional
$\psi : \mathcal M \to \mathbb R^k$ is \emph{pathwise differentiable}
at $P$ relative to a tangent space $T$ if there exists a
continuous linear map $\dot\psi : T \to \mathbb R^k$ such that for
every QMD submodel $t \mapsto P_t$ through $P$ with score $g \in T$,
the difference quotient $t^{-1}(\psi(P_t) - \psi(P))$ converges to
$\dot\psi(g)$ as $t \to 0$.
\begin{leanbox}
\small
\begin{verbatim}
structure PathwiseDifferentiableAt_vec
    (ψ : Measure Ω → EuclideanSpace ℝ (Fin k)) where
  derivative : T →L[ℝ] EuclideanSpace ℝ (Fin k)
  derivative_spec :
    ∀ (γ : QMDPath.QMDPath P),
      ∀ (h_in_T : (γ.score : ↥(L2ZeroMean P)) ∈ T),
        Filter.Tendsto
          (fun t : ℝ => t⁻¹ • (ψ (γ.curve t) - ψ P))
          (nhdsWithin 0 {0}ᶜ) (nhds (derivative ⟨γ.score, h_in_T⟩))
\end{verbatim}
\end{leanbox}
\noindent Both fields match the source definition: the source definition
packs (existence of a continuous-linear derivative) and
(convergence along every QMD submodel) and we transcribe them as
the two fields.  Notably \textbf{absent}: uniform-Hadamard control,
or any explicit remainder rate.  \cite[Chapter~25.3]{vanderVaart1998} phrases convergence
pointwise in $t \to 0$ along each fixed submodel; uniformity over
submodels and remainder rates appear later (Chapter~25.6, for specific
applications) and are not part of the definition.  The rejected
\verb|hadamard_remainder| field of Section~\ref{sec:eval-semipar-convolution-drift}
would have added such a rate condition here, strictly strengthening
the source's pointwise statement.

\noindent \textbf{Parametric family (\texttt{ParametricFamily}, \texttt{IsPDFOf}).}
\cite[Chapter~7]{vanderVaart1998}, opening pages.  Carries the
per-parameter density, its measurability, and nonnegativity; the
normalization and integrability are bundled in the companion predicate
\verb|IsPDFOf|.
\begin{leanbox}
\small
\begin{verbatim}
structure ParametricFamily
    (𝓧 : Type*) [MeasurableSpace 𝓧] (Θ : Type*) where
  density : Θ → 𝓧 → ℝ
  density_meas : ∀ θ, Measurable (density θ)
  density_nonneg : ∀ θ x, 0 ≤ density θ x

structure IsPDFOf
    {𝓧 : Type*} [MeasurableSpace 𝓧] {Θ : Type*}
    (M : ParametricFamily 𝓧 Θ) (μ : Measure 𝓧) : Prop where
  density_integral_eq_one : ∀ θ, ∫ x, M.density θ x ∂μ = 1
  density_integrable : ∀ θ, Integrable (M.density θ) μ
\end{verbatim}
\end{leanbox}

\noindent \textbf{Score function and Fisher information (\texttt{ScoreFunction}, \texttt{fisherInformation}).}
\cite[Theorem~7.2]{vanderVaart1998}, p.~94.  The score is a
measurable map; the $L^2$ properties (zero mean, finite Fisher
information) are downstream consequences of DQM, not structure
fields (see Section~\ref{sec:concept-layer}).
\begin{leanbox}
\small
\begin{verbatim}
structure ScoreFunction
    (M : ParametricFamily 𝓧 Θ) (θ : Θ) where
  toFun : 𝓧 → Θ
  measurable : Measurable toFun

noncomputable def fisherInformation
    (M : ParametricFamily 𝓧 Θ) (μ : Measure 𝓧) (θ : Θ) (ℓ : 𝓧 → Θ) :
    Θ → Θ → ℝ :=
  fun u v => ∫ x, (⟪u, ℓ x⟫ * ⟪v, ℓ x⟫) * M.density θ x ∂μ
\end{verbatim}
\end{leanbox}

\noindent \textbf{Product measure $P^n_\theta$ (\texttt{productMeasure}).}
\cite[Chapter~7]{vanderVaart1998}, the $n$-fold i.i.d.\ product
$P^n_\theta$ used throughout the parametric cornerstones.  Wraps
Mathlib's \verb|Measure.pi| specialized to a parametric-family
density.
\begin{leanbox}
\small
\begin{verbatim}
noncomputable def productMeasure
    (M : ParametricFamily 𝓧 (Θ k)) (μ : Measure 𝓧) (θ : Θ k) (n : ℕ) :
    Measure (Fin n → 𝓧) :=
  Measure.pi (fun _ =>
    μ.withDensity fun x => ENNReal.ofReal (M.density θ x))
\end{verbatim}
\end{leanbox}

\noindent \textbf{Weak convergence (\texttt{WeakConverges}).}
A predicate-form wrapper for
\cite{vanderVaart1998}'s $\rightsquigarrow$.  Mathlib already
carries weak convergence as the weak topology on
\texttt{ProbabilityMeasure} (with the bounded-continuous-test-function
characterization as the
\texttt{tendsto\_iff\_\allowbreak forall\_\allowbreak integral\_\allowbreak tendsto}
theorem); the project's
\verb|WeakConverges| restates the same content as a direct
predicate on raw \verb|Measure E| arguments, so that cornerstone
signatures can pass raw measure sequences directly.
\begin{leanbox}
\small
\begin{verbatim}
def WeakConverges {E : Type*} [MeasurableSpace E] [TopologicalSpace E]
    (μ : ℕ → Measure E) (ν : Measure E) : Prop :=
  ∀ f : E →ᵇ ℝ,
    Tendsto (fun n => ∫ x, f x ∂(μ n)) atTop
      (𝓝 (∫ x, f x ∂ν))
\end{verbatim}
\end{leanbox}

\noindent \textbf{Bowl-shaped loss (\texttt{BowlShaped}).}
\cite[Chapter~8.4]{vanderVaart1998}, p.~113.  Symmetric, convex
sublevel sets, plus a measurability adapter (source-implicit,
required in Lean for the integral on the LAM bound's LHS to be
defined).
\begin{leanbox}
\small
\begin{verbatim}
structure BowlShaped [AddCommGroup E] [Module ℝ E] [MeasurableSpace E]
    (L : E → ℝ≥0∞) : Prop where
  measurable : Measurable L
  symm : ∀ x, L (-x) = L x
  convex_sublevel : ∀ c : ℝ≥0∞, Convex ℝ {x | L x ≤ c}
\end{verbatim}
\end{leanbox}

\noindent \textbf{Local asymptotic L-risk (\texttt{localAsymptoticRisk}).}
\cite[Theorem~8.11]{vanderVaart1998}, p.~118, the LHS of the LAM
bound: worst-case asymptotic L-risk over local finite alternatives.
\begin{leanbox}
\small
\begin{verbatim}
noncomputable def localAsymptoticRisk
    (M : ParametricFamily 𝓧 (Θ k)) (μ : Measure 𝓧) (θ₀ : Θ k)
    (T : ∀ n, (Fin n → 𝓧) → 𝓨 d) (ψ : Θ k → 𝓨 d)
    (L : 𝓨 d → ℝ≥0∞) : ℝ≥0∞ :=
  ⨆ I : Finset (Θ k), Filter.liminf
    (fun n : ℕ => ⨆ h ∈ I,
      ∫⁻ ω, L ((Real.sqrt n) • (T n ω - ψ (θ₀ + (Real.sqrt n)⁻¹ • h)))
            ∂(AsymptoticRepresentation.productMeasure M μ (θ₀ + (Real.sqrt n)⁻¹ • h) n))
    Filter.atTop
\end{verbatim}
\end{leanbox}

\noindent \textbf{QMD submodel (\texttt{QMDPath}).}
\cite[Chapter~25.3, Eq.~(25.13)]{vanderVaart1998}, pp.~362--363.
A quadratic-mean-differentiable submodel through $P$ in the
\textbf{dominated} specialization authorized by the footnote on
p.~362; the score lives in the closed mean-zero $L^2$
subspace \verb|L2ZeroMean P| (defined below).
\begin{leanbox}
\small
\begin{verbatim}
structure QMDPath (P : Measure Ω) [IsProbabilityMeasure P] where
  curve : ℝ → Measure Ω
  curve_at_zero : curve 0 = P
  curve_isProbability : ∀ t, IsProbabilityMeasure (curve t)
  dominating : Measure Ω
  curve_absContinuous : ∀ t, curve t ≪ dominating
  dominating_sigmaFinite : SigmaFinite dominating
  score : ↥(L2ZeroMean P)
  qmd_limit :
    Tendsto
      (fun t : ℝ =>
        eLpNorm (fun ω : Ω =>
          Real.sqrt ((curve t).rnDeriv dominating ω).toReal
            - Real.sqrt ((curve 0).rnDeriv dominating ω).toReal
            - (t / 2) * (score : Ω → ℝ) ω
                * Real.sqrt ((curve 0).rnDeriv dominating ω).toReal)
          2 dominating / ENNReal.ofReal |t|)
      (𝓝[≠] 0) (𝓝 (0 : ℝ≥0∞))
\end{verbatim}
\end{leanbox}

\noindent \textbf{Zero-mean $L^2$ subspace (\texttt{L2ZeroMean}).}
The closed submodule of \verb|Lp ℝ 2 P| of mean-zero functions:
the kernel of the integral functional, restricted to $L^2(P)$.  The
semi-parametric tangent space is a submodule of this object.
\begin{leanbox}
\small
\begin{verbatim}
noncomputable def L2ZeroMean (P : Measure Ω) [IsFiniteMeasure P] :
    Submodule ℝ (Lp ℝ 2 P) :=
  LinearMap.ker (integralL2 P).toLinearMap
\end{verbatim}
\end{leanbox}

\noindent \textbf{Efficient influence function, vector form (\texttt{IsEfficientInfluenceFunction\_vec}).}
\cite[Chapter~25.3]{vanderVaart1998}, pp.~363--365.
\verb|IsEfficientInfluenceFunction_vec| lifts the coordinate-wise
efficient-influence-function predicate to a $k$-tuple.  The
semi-parametric Cram\'er--Rao bound is the $k \times k$ Gram matrix
of this tuple --- \cite[p.~365]{vanderVaart1998}'s optimal asymptotic
covariance $P\tilde\psi_P\tilde\psi_P^{\top}$, with entries
$\langle \mathrm{IF}_i, \mathrm{IF}_j\rangle_{L^2(P)}$.  The cornerstone
signatures take this matrix as Mathlib's \verb|Matrix.gram| applied to
the tuple of efficient influence functions, so no project-level wrapper
is introduced.
\begin{leanbox}
\small
\begin{verbatim}
def IsEfficientInfluenceFunction_vec
    (Dψ : T →L[ℝ] EuclideanSpace ℝ (Fin k))
    (IF : Fin k → ↥(L2ZeroMean P)) : Prop :=
  ∀ i, IsEfficientInfluenceFunction P T
    (EuclideanSpace.proj i ∘L Dψ) (IF i)
\end{verbatim}
\end{leanbox}

\noindent \textbf{Regular estimator, semi-parametric form (\texttt{IsRegularEstimator\_vec}).}
\cite[Chapter~25.3]{vanderVaart1998}, pp.~365--366; the
semi-parametric analog of \verb|RegularEstimatorSequence|. The source
phrases regularity by selecting, for each score direction
$g \in \dot{\mathcal{P}}_P$, a submodel (``write $P_{t,g}$ for a
submodel''); the Lean definition records this submodel as an
existentially quantified \verb|chosenFamily| with score $g$, along which
the rescaled, perturbed-truth--recentered estimator converges weakly to
the common limit law $L$.  The score direction is quantified over the
algebraic span of \verb|T_set.carrier|
(\cite[p.~366]{vanderVaart1998}, ``for every $g \in \mathrm{lin}\, g_p$'');
the pathwise derivative \verb|hψ| and the efficient influence function
\verb|hEIF| are quantified over the L\textsuperscript{2}-closed span
\verb|tangentSpace T_set|.
\begin{leanbox}
\small
\begin{verbatim}
def IsRegularEstimator_vec
    (P : Measure Ω) [IsProbabilityMeasure P]
    (T_set : TangentSpec P)
    {k : ℕ}
    (ψ : Measure Ω → EuclideanSpace ℝ (Fin k))
    {IF_eff : Fin k → ↥(L2ZeroMean P)}
    (hψ : PathwiseDifferentiableAt_vec P (tangentSpace T_set) ψ)
    (_hEIF : IsEfficientInfluenceFunction_vec
              (P := P) (T := tangentSpace T_set)
              hψ.derivative IF_eff)
    (T_n : ∀ n, (Fin n → Ω) → EuclideanSpace ℝ (Fin k))
    (L : Measure (EuclideanSpace ℝ (Fin k)))
    [IsProbabilityMeasure L] :
    Prop :=
  ∃ chosenFamily :
      ∀ (g : ↥(L2ZeroMean P)),
        (g : ↥(L2ZeroMean P)) ∈
          Submodule.span ℝ T_set.carrier → QMDPath P,
    (∀ (g : ↥(L2ZeroMean P))
        (hg : (g : ↥(L2ZeroMean P)) ∈
          Submodule.span ℝ T_set.carrier),
      (chosenFamily g hg).score = g) ∧
    (∀ (g : ↥(L2ZeroMean P))
        (hg : (g : ↥(L2ZeroMean P)) ∈
          Submodule.span ℝ T_set.carrier),
      WeakConverges
        (fun n : ℕ =>
          (MeasureTheory.Measure.pi
              (fun _ : Fin n =>
                (chosenFamily g hg).curve
                  ((Real.sqrt n)⁻¹))).map
            (fun X : Fin n → Ω =>
              Real.sqrt n •
                (T_n n X
                  - ψ ((chosenFamily g hg).curve
                      ((Real.sqrt n)⁻¹)))))
        L)
\end{verbatim}
\end{leanbox}

\end{document}